\documentclass[11pt,a4paper]{article}
\usepackage[hyperref]{acl2021}
\usepackage{times}
\usepackage{latexsym}
\usepackage{url}
\usepackage{float,cuted}
\usepackage{graphicx}
\usepackage{subfigure}
\usepackage{amssymb}
\usepackage{amsmath,bm}
\usepackage{paralist,tabularx,multirow}
\usepackage{caption}
\usepackage{booktabs}

\usepackage{microtype}

\aclfinalcopy 

\title{RealTranS: End-to-End Simultaneous Speech Translation with Convolutional Weighted-Shrinking Transformer}

\author{Xingshan Zeng, Liangyou Li, Qun Liu \\
  Huawei Noah's Ark Lab \\
  \texttt{\{zeng.xingshan,liliangyou,qun.liu\}@huawei.com} \\}

\date{}

\begin{document}
\maketitle

\begin{abstract}
End-to-end simultaneous speech translation (SST), which directly translates speech in one language into text in another language in real-time, is useful in many scenarios but has not been fully investigated. In this work, we propose RealTranS, an end-to-end model for SST. To bridge the modality gap between speech and text, RealTranS gradually downsamples the input speech with interleaved convolution and unidirectional Transformer layers for acoustic modeling, and then maps speech features into text space with a weighted-shrinking operation and a semantic encoder. Besides, to improve the model performance in simultaneous scenarios, we propose a blank penalty to enhance the shrinking quality and a Wait-K-Stride-N strategy to allow local reranking during decoding.
Experiments on public and widely-used datasets show that RealTranS with the Wait-K-Stride-N strategy outperforms prior end-to-end models as well as cascaded models in diverse latency settings.
\end{abstract}
\section{Introduction}
\begin{figure}[t]
\centering
\subfigure[Cascaded Model]{
\includegraphics[width=0.225\textwidth]{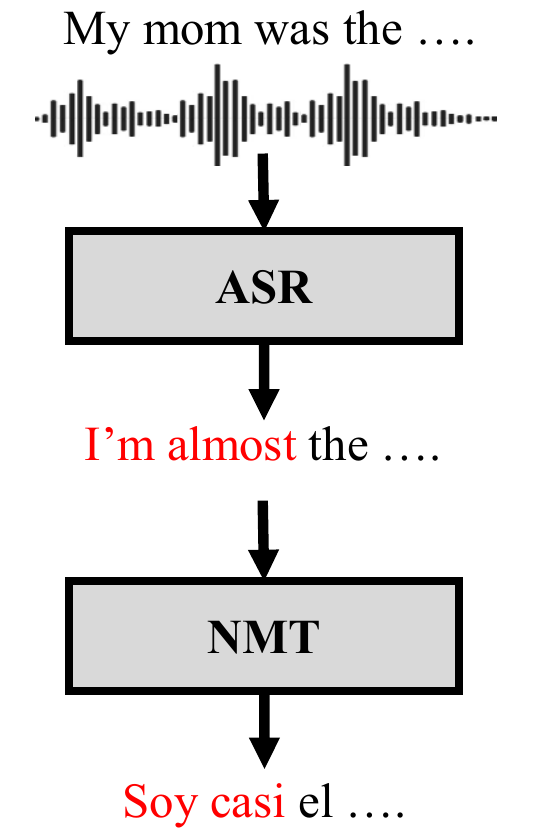}
}
\subfigure[RealTranS]{
\includegraphics[width=0.225\textwidth]{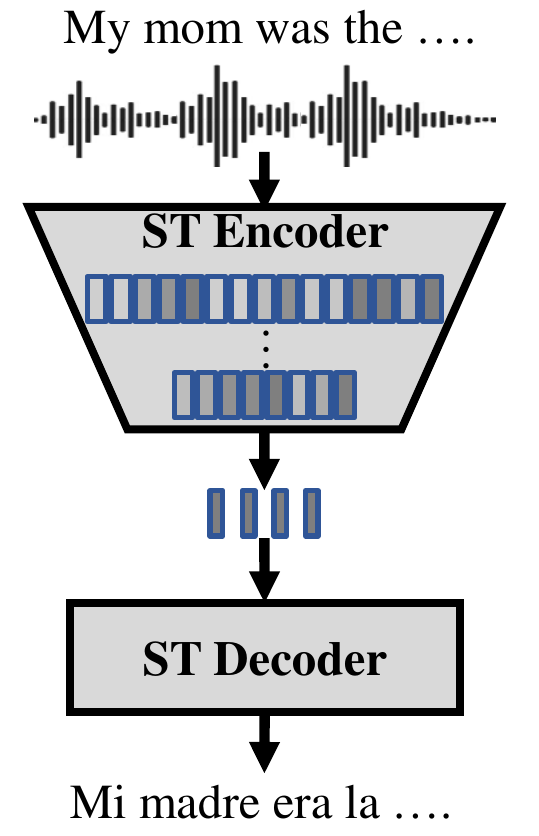}
}
\caption{
An example for a cascaded model and our RealTranS. The ASR part in the cascaded model wrongly recognizes ``My mom was'' as ``I'm almost''. The error is propagated to NMT and leads to a wrong translation. RealTranS avoids such errors and translates accurately.
}
\label{fig:intro}
\end{figure}

Simultaneous speech translation (SST)~\cite{DBLP:journals/mt/FugenWK07,oda-etal-2014-optimizing,ren-etal-2020-simulspeech} aims to translate speech in one language into text in another language concurrently. It is useful in many scenarios, like synchronous interpretation in international conferences, automatic caption for live videos, etc. However, prior studies either focus on full sentence speech translation (ST)~\cite{DBLP:journals/corr/BerardPSB16,DBLP:journals/corr/WeissCJWC17,DBLP:conf/interspeech/LiuXZHWWZ19} or simultaneous text-to-text machine translation (STT)~\cite{DBLP:journals/corr/ChoE16,gu-etal-2017-learning,dalvi-etal-2018-incremental} which takes a segmented output from an automatic speech recognition (ASR) system as input. Such two-stage models (i.e., cascaded models) inevitably introduce error propagation and also increase translation latency (see Figure~\ref{fig:intro}). 
\citet{ren-etal-2020-simulspeech} propose an end-to-end SST system called SimulSpeech, but they ignore the modality gap between speech and text, which is important for improving translation quality~\cite{DBLP:journals/corr/abs-2010-14920}.

In this paper, we propose RealTranS model for SST. To relieve the burden of our encoder~\cite{wang-etal-2020-curriculum,DBLP:journals/corr/abs-2010-14920}, we decouple it into three parts: acoustic encoder, weighted-shrinking operation, and semantic encoder. We apply Conv-Transformer~\cite{DBLP:conf/interspeech/HuangHYC20} as our acoustic encoder, which gradually downsamples the input speech and learns acoustic information with interleaved convolution and Transformer layers.
The weighted-shrinking operation bridges the length gap between speech and text, by weighted summing up the frames in one detected segment based on the posterior probabilities generated by a CTC module~\cite{DBLP:conf/icml/GravesFGS06}.
Finally, we use a semantic encoder to extract semantic features and deliver them to the decoder for translation. 

To enable simultaneous decoding, unidirectional Transformer is applied in our encoder. This inevitably affects the performance of the CTC module and the following shrinking operation. To alleviate this, we introduce a blank-limited CTC loss, which adds a blank penalty to the traditional CTC loss to encourage the model to produce non-blank labels, given the observation that CTC tends to produce peaky
distribution as a kind of overfitting~\cite{DBLP:conf/nips/LiuJZ18} by over predicting blank labels. Accordingly, the shrinking quality can be improved. Furthermore, we propose a new simultaneous strategy Wait-K-Stride-N which allows local reranking during decoding. This strategy can resolve the inherent drawback of the conventional Wait-K strategy~\cite{ma-etal-2019-stacl}, which cannot apply vanilla beam search efficiently~\cite{zheng-etal-2019-speculative}.

Experiments on Augmented LibriSpeech En--Fr, MUST-C En--Es and En--De datasets demonstrate the effectiveness of the Wait-K-Stride-N strategy, and show that RealTranS achieves better performance than the prior end-to-end model SimulSpeech~\cite{ren-etal-2020-simulspeech} as well as the cascaded models. Further analysis and ablation study reveal the effects of our proposed modules in RealTranS. We also compare RealTranS with other methods on full sentence ST. Results show that RealTranS achieves competitive or even better results, indicating its superiority.

In summary, the contributions of this work include the following aspects: 

\begin{itemize}
\item We propose RealTranS for SST, which can gradually bridge the modality gap between speech and text with the help of gradual downsampling and weighted shrinking. 

\item We introduce a blank penalty and the Wait-K-Stride-N strategy to improve the performance in simultaneous translation scenarios. 

\item Extensive experiments on public and widely-used datasets show the superiority of our RealTranS model and our Wait-K-Stride-N strategy in diverse latency settings.
\end{itemize}

\section{Related Work}
\paragraph{Speech Translation.}
Speech translation (ST) has recently attracted intensive attention from the AI community.
Earlier works are mostly based on cascaded models, which perform NMT on the outputs of ASR systems~\cite{DBLP:conf/icassp/Ney99,DBLP:conf/icassp/MathiasB06,sperber-etal-2017-neural,DBLP:conf/slt/BaharBSN21}. Cascaded models inevitably introduce error propagation from ASR~\cite{DBLP:journals/corr/WeissCJWC17}. To avoid this problem and for better efficiency, end-to-end ST models are proposed and become popular in recent years~\cite{DBLP:journals/corr/BerardPSB16,DBLP:conf/icassp/BerardBKP18,DBLP:conf/interspeech/BansalKLLG18,DBLP:conf/interspeech/GangiNT19}. 
To alleviate the data scarcity problem of end-to-end ST models, various techniques are utilized, including pre-training~\cite{bansal-etal-2019-pre}, multi-task learning~\cite{anastasopoulos-chiang-2018-tied}, knowledge distillation~\cite{DBLP:conf/interspeech/LiuXZHWWZ19,ren-etal-2020-simulspeech}, data synthesis~\cite{DBLP:conf/icassp/JiaJMWCCALW19}, self-supervised learning~\cite{DBLP:journals/corr/abs-2010-11445} and speech augmentation techniques like SpecAugment~\cite{DBLP:journals/corr/abs-1911-08876} or speed perturbation~\cite{DBLP:conf/icassp/StoianBG20}.

Some studies focus on how to bridge the gap between different modalities (speech and text) or different modules (acoustic and semantic modeling). \citet{DBLP:conf/aaai/WangWLY020,wang-etal-2020-curriculum} propose a TCEN model and a curriculum pre-training technique to make sure the modules learn desired information, respectively. \citet{salesky-black-2020-phone} explore phone features as intermediate representations to improve performance, while \citet{DBLP:journals/corr/abs-2009-09704} use pre-trained BERT to guide the model to learn semantic knowledge. Modality Agnostic Meta-Leaning is also exploited for ST in \citet{DBLP:journals/corr/abs-1911-04283}. To bridge the length gap, \citet{zhang-etal-2020-adaptive} propose adaptive feature selection, while \citet{DBLP:journals/corr/abs-2009-09737} and \citet{DBLP:journals/corr/abs-2010-14920} exploit the CTC-based~\cite{DBLP:conf/icml/GravesFGS06} shrinking mechanism. 
Nevertheless, they do not explore in simultaneous scenarios, where encoding quality inevitably suffers because of lacking future information in unidirectional encoders.

\begin{figure*}[t]
\centering
\includegraphics[width=158mm]{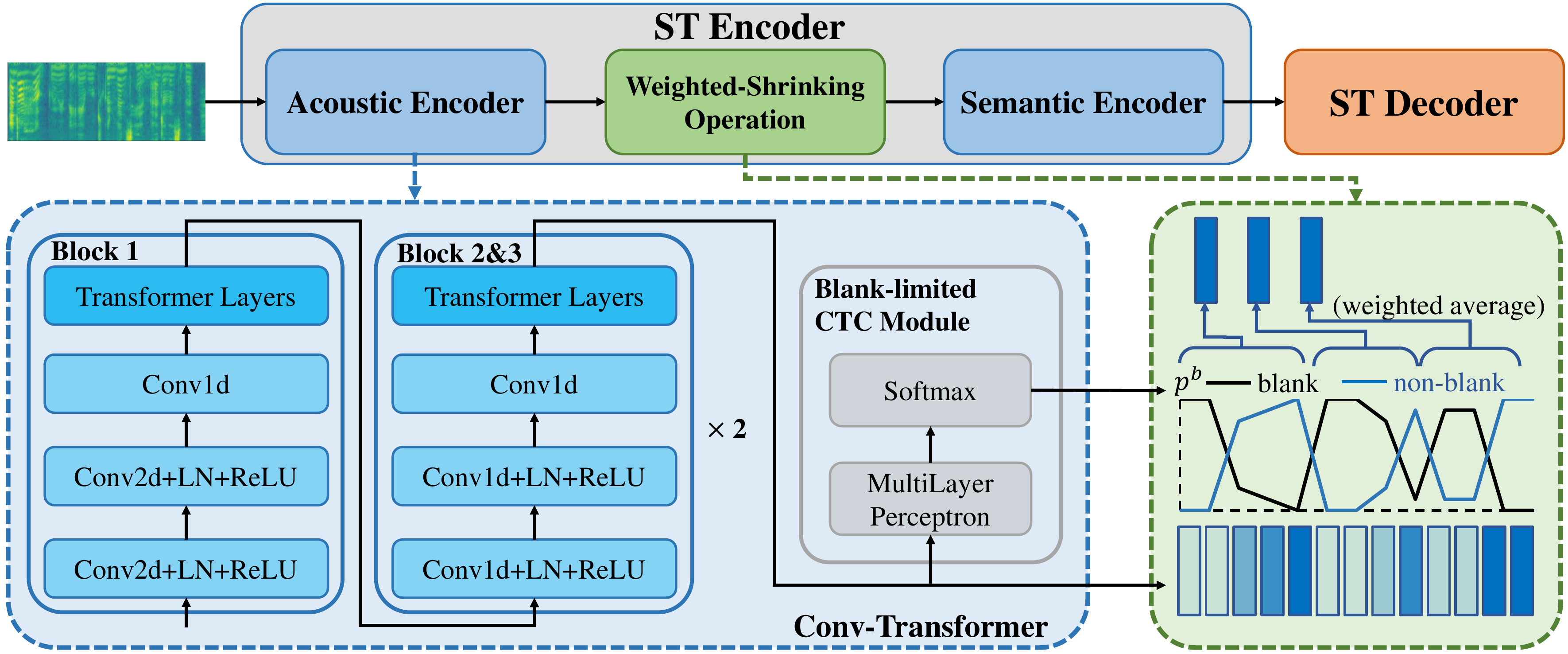}
\caption{
Overall structure of the proposed RealTranS model. 
}
\label{fig:model}
\end{figure*}
\paragraph{Simultaneous Translation.}
Previous studies on simultaneous translation focus on text-to-text scenarios (STT)~\cite{DBLP:journals/corr/ChoE16,gu-etal-2017-learning,dalvi-etal-2018-incremental}, where fixed policies~\cite{ma-etal-2019-stacl} and adaptive policies~\cite{arivazhagan-etal-2019-monotonic,zheng-etal-2019-simpler,zheng-etal-2019-simultaneous} are proposed to decide when to read and write tokens. \cite{ma-etal-2019-stacl} proposed a simple yet effective strategy, Wait-K, based on a prefix-to-prefix framework. It first waits for the first k tokens, and then start to generate target tokens concurrently with the source stream. It achieves competitive performance in simultaneous translation~\cite{zheng-etal-2019-simpler,zheng-etal-2020-simultaneous}.

Traditional simultaneous speech-to-text translation (SST) mainly depends on the ASR segmentation and then performs NMT based on the streaming segmented chunks~\cite{oda-etal-2014-optimizing,iranzo-sanchez-etal-2020-direct}. There is little attention on end-to-end SST. \citet{ren-etal-2020-simulspeech}, to our knowledge, first propose an end-to-end model called SimulSpeech with multi-task learning and knowledge distillation, and apply the Wait-K strategy to perform simultaneous translation. \citet{ma-etal-2020-simulmt} explore how to define a ``token" in source speech, and then adapt methods from STT to SST. And \citet{DBLP:journals/corr/abs-2011-00033} introduce a memory-augmented Transformer to tackle the streaming speech input. However, none of them investigate 
the modality gap between speech and text.

\section{The RealTranS Model}
Our RealTranS follows the sequence-to-sequence architecture, which consists of an ST encoder and an ST decoder. The ST encoder is decoupled into three parts, an acoustic encoder, a weighted shrinking operation, and a semantic encoder, to gradually map speech inputs into semantic representation space of text. Figure~\ref{fig:model} shows the architecture. 

\subsection{Problem Formulation} \label{sec:mod-pf}
Speech translation corpora usually contain triples of speech, transcription and translation, denoted as $\mathcal{D}_{ST} = \{(\bm{x}, \bm{z}, \bm{y})\}$. Specifically, $\bm{x} = (x_1, x_2, ..., x_{T_x})$ is a sequence of speech features extracted from speech signals, e.g., filterbanks. $\bm{z} = (z_1, z_2, ..., z_{T_z})$ and $\bm{y} = (y_1, y_2, ..., y_{T_y})$ are the corresponding transcription in source language and translation in target language. $T_x$, $T_z$, and $T_y$ are the lengths of speech, transcription and translation, respectively, where usually $T_x \gg T_z$ and $T_x \gg T_y$. A typical end-to-end model only makes use of $\bm{x}$ and $\bm{y}$, while $\bm{z}$ can be used as multi-task training for other objectives, like CTC loss.

\subsection{Acoustic Encoder} \label{sec:mod-ae}
Acoustic encoder mainly encodes speech features $\bm{x}$ into a hidden space to learn acoustic knowledge. We apply Conv-Transformer~\cite{DBLP:conf/interspeech/HuangHYC20} to extract the desired features. It contains three blocks, each of which is composed of three convolution layers followed by unidirectional Transformer layers (see lower left in Figure~\ref{fig:model}) to prevent from leveraging future context in SST. Similar to \citet{DBLP:conf/interspeech/HuangHYC20}, we make the model aware of limited future frames with a look-ahead window in the convolution layers, to help improve acoustic modeling.
At the same time, we gradually downsample the long speech features by setting the stride size to 2 in the second convolution layer in each block. In this way, the speech features are gradually reduced and approach the length of the corresponding text.

\begin{figure}[t]
\centering
\includegraphics[width=68mm]{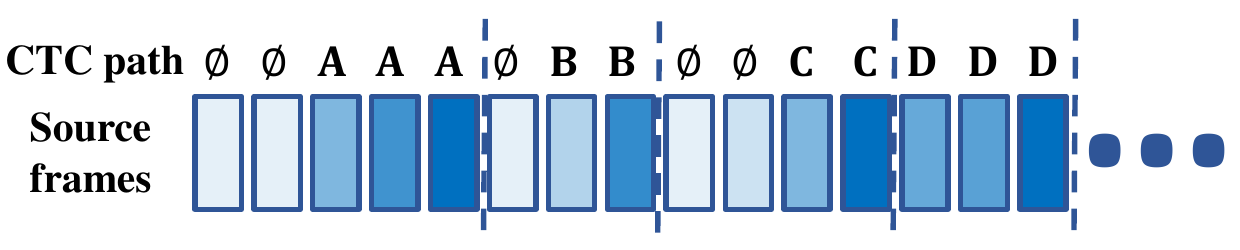}
\caption{
An example of how to define the word boundaries. The dash lines are the boundaries.
}
\label{fig:boundaries}
\end{figure}

To predict the word boundaries\footnote{For ease of understanding, we use word boundaries here. Same methods can be applied to boundaries between phones, chars, subwords, etc., depending on the the granularity of the CTC loss units. We use subword units in our experiments.} in the speech input and further improve the learned acoustic features, we adopt the Connectionist Temporal Classification (CTC) \cite{DBLP:conf/icml/GravesFGS06} module on top of the acoustic encoder. This module contains a Multilayer Perceptron (MLP) followed by a Softmax operator. CTC predicts a path $\bm{\pi} = (\pi_1, \pi_2, ..., \pi_{T^\prime_x})$, where $T^\prime_x$ is the length of hidden states after the Conv-Transformer. And $\pi_t \in \mathcal{V} \cup \{\phi\}$ can be either a  token in the source vocabulary $\mathcal{V}$ or the blank symbol $\phi$. CTC paths have the many-to-one mapping to output sequences by removing
blank symbols and consecutively repeated labels, denoted as operation $\mathcal{B}$. Therefore, CTC loss is defined as follows:
\begin{equation} \label{eq:ctc-loss} \small
    \mathcal{L}_{CTC} = - \sum_{(\bm{x},\bm{z}) \in \mathcal{D}} \, \sum_{\bm{\pi} \in \mathcal{B}^{-1}(\bm{z})} p(\bm{\pi} | \bm{x})
\end{equation}
where $\mathcal{B}^{-1}(\bm{z})$ denotes all possible CTC paths that can be mapped to the transcription $\bm{z}$. With the CTC module, we can define a word boundary between two frames where the first frame has a non-blank label while the second frame has a different label from the first one. Figure \ref{fig:boundaries} shows an example of word boundaries.

Due to the data scarcity problem in ST and CTC's inherent  characteristics~\cite{DBLP:conf/nips/LiuJZ18}, the detected word boundaries are usually not accurate enough. The module will overly predict the occurrence of blank labels (as a kind of overfitting),
resulting in a large gap between the number of detected boundaries and the number of tokens in transcription,
especially when unidirectional Transformer is applied (see Table~\ref{tab:shrink-quality}). To alleviate the problem, we add a blank penalty to encourage the module to produce non-blank labels. It is called blank-limited CTC loss and defined as follows:
\begin{equation} \label{eq:bp-ctc-loss} \small
    \mathcal{L}^{\prime}_{CTC} = \mathcal{L}_{CTC} + \lambda \sum_{\bm{x} \in \mathcal{D}} \, \sum_{\pi_t \in \bm{\pi}(\bm{x})} p(\pi_t = \phi | \bm{x})
\end{equation}
where $\bm{\pi}(\bm{x})$ means the argmax results of CTC softmax outputs. $\lambda$ controls the effect of blank penalty.

\subsection{Weighted-Shrinking Operation} \label{sec:mod-wo}
The length gap between acoustic features and the corresponding transcription and translation is still large after the gradual downsampling in Section~\ref{sec:mod-ae}. Inspired by previous studies~\cite{DBLP:journals/corr/abs-2009-09737,DBLP:journals/corr/abs-2010-14920}, we adopt a shrinking operation to bridge the gap based on CTC predictions. Prior works usually either remove blank frames and then average repeated frames~\cite{DBLP:journals/corr/abs-2009-09737} or select a single representative frame~\cite{DBLP:journals/corr/abs-2010-14920} in a detected segment (i.e., frames between two word boundaries). However, there might be useful information in the detected blank or repeated frames, especially when the boundary detection is not accurate enough (see Section~\ref{sec:mod-ae}). 

We propose a weighted-shrinking mechanism to tackle the problem. We assume that the probability of a frame to be labeled as ``blank'' represents the confidence that the model ``thinks'' it is not important. Therefore, for the frames in one segment, their weights are decided by the probabilities to be blank labels. The representation of the segment will be the weighted average of the corresponding frames. The specific operation is shown as follows:
\begin{equation} \label{eq:we-shrink} \small
    \bm{h}_{t^{\prime}} = \sum_{t \in seg\,t^{\prime} } \bm{h}_t\frac{\exp(\mu (1 - p_{t}^{b}))}{\sum_{s\in seg_{t^\prime}}\exp(\mu (1 - p_{s}^{b}))}
\end{equation}
where $p_{t}^{b}$ denotes the probability of the frame $t$ to be blank, and $\bm{h}_t$ represents the hidden state of the frame $t$ in our acoustic encoder. $\mu \ge 0$ controls the temperature of the distribution (i.e., Softmax function). When $\mu = 0$, it means that we simply average the frames; and when $\mu \to \infty$ it degenerates to the general shrinking mechanism where only the representative frames with the highest confidence are selected.

\subsection{Semantic Encoder}
\label{sec:mod-se}
The shrinking operation only bridges the length gap between speech and text, but the shrunk representations still lack semantic information. Therefore, we apply a semantic encoder on top of the shrunk representations \cite{DBLP:journals/corr/abs-2010-14920}. 
It first applies a positional embedding layer, and then follows several Transformer layers (also unidirectional to mask future context), to extract semantic representations.

\subsection{ST Decoder}
\label{sec:mod-sd}
A similar decoder as the basic Transformer architecture in NMT is adopted, where several Transformer decoder layers are stacked on top of target embeddings. To simulate simultaneous translation, we follow the prefix-to-prefix framework~\cite{ma-etal-2019-stacl} and mask certain future context in cross attention to ensure that the model predicts the current token based on only part of the input from the ST encoder. How much context the model can see depends on the simultaneous strategy that is applied. For example, a Wait-K~\cite{ma-etal-2019-stacl} based decoder predicts the $t$-th token based on the first $t+k-1$ hidden states produced by the ST encoder.

\begin{figure}[t]
\centering
\includegraphics[width=68mm]{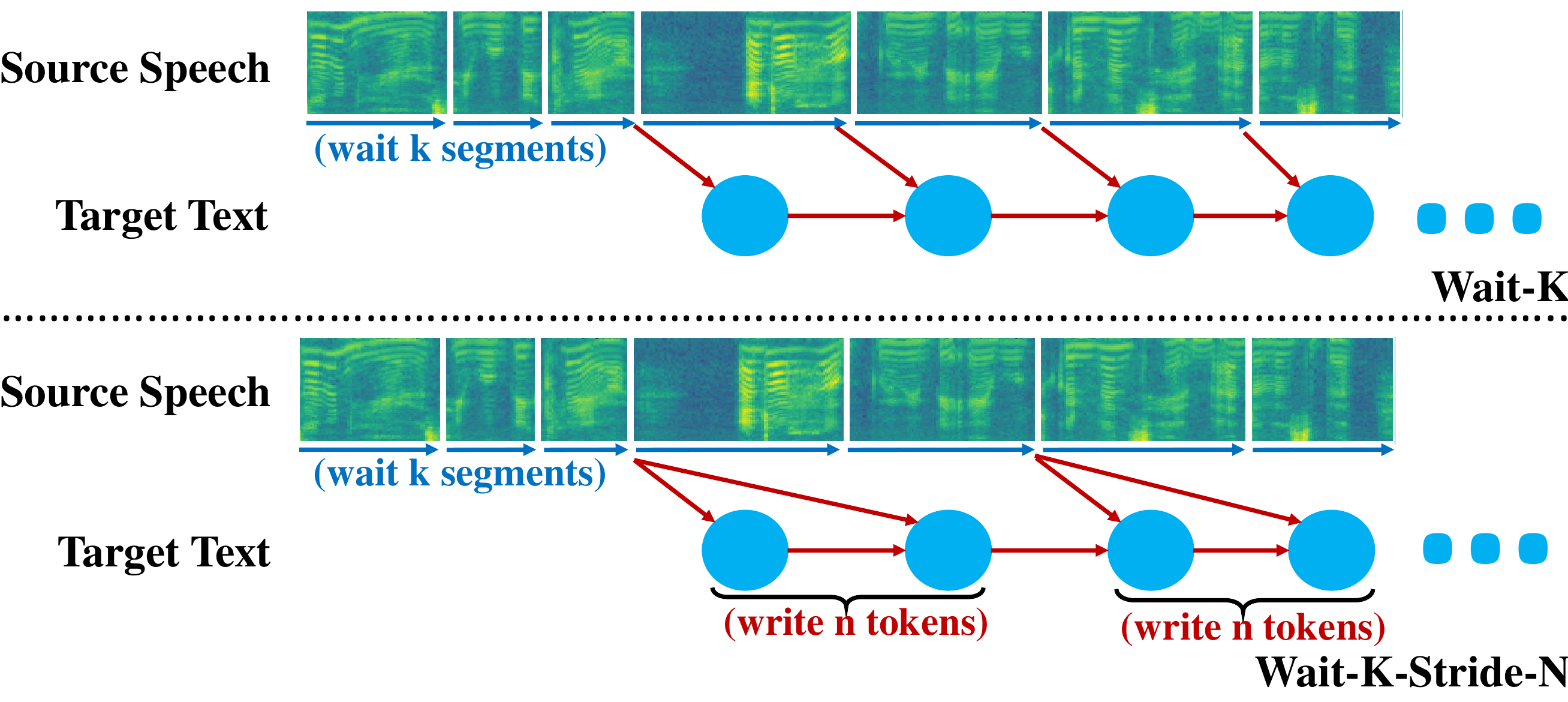}
\caption{
Wait-K vs. Wait-K-Stride-N (in this example, we set $k=3$ and $n=2$). 
}
\label{fig:strategy}
\end{figure}
\paragraph{Wait-K-Stride-N.}
There is one drawback in the conventional Wait-K strategy -- it cannot perform vanilla beam search while decoding except for the long-tail scenario~\cite{ma-etal-2019-stacl,zheng-etal-2019-speculative}, though beam search has been proven very effective in improving translation quality.
Based on the prefix-to-prefix framework and STATIC-RW strategy proposed in \citet{dalvi-etal-2018-incremental}, we propose the Wait-K-Stride-N strategy to allow using beam search for local reranking during simultaneous decoding. Similar to the Wait-K strategy, our strategy first reads $k$ input units (tokens in MT or segments in ST). 
Then, the model repeatedly performs $n$ write and read operations until the end of the sentence (see Figure~\ref{fig:strategy}). In this way, the translation latency is close to Wait-K, but we can perform beam search during the $n$ write operations.
The objective with such a strategy is hence defined as follows:
\begin{equation} \label{eq:st-loss} \small
    \mathcal{L}_{ST} = - \sum_{(\bm{x},\bm{y}) \in \mathcal{D}} \, \prod_{t=1}^{T_y} p(y_t | y_{<t},x_{\le n\lfloor(t-1)/n\rfloor+k})
\end{equation}
where $y_{<t}$ represents the target tokens before $y_t$, $T_y$ is the length of the target sentence, and $x_{\le t}$ represents the first $t$ detected source segments.

\subsection{Training Procedure}
The total objective of our model will be the sum of the CTC part and the ST part:
\begin{equation} \label{eq:loss} \small
    \mathcal{L} = \mathcal{L}_{ST} + \alpha \mathcal{L}^{\prime}_{CTC}
\end{equation}
where $\alpha$ controls the influence of the CTC part.

To enhance the CTC quality, we also apply a pre-training procedure~\cite{DBLP:conf/icassp/StoianBG20}. We only use CTC loss to pre-train the acoustic encoder\footnote{We do not pre-train the decoder for simplicity though it might further improve our performance.}. In this way, we can prevent the training waste~\cite{DBLP:conf/aaai/WangWLY020}, and focus on improving the alignment results, which are essential for shrinking operations (see Table~\ref{tab:shrink-quality}). The whole model is then fine-tuned with the whole ST corpus.

\section{Experimental Setup}
\subsection{Datasets}
We conduct experiments on three publicly available datasets:
Augmented LibriSpeech English-French (En--Fr) corpus~\cite{DBLP:conf/lrec/KocabiyikogluBK18}, and 
MUST-C English-Spanish (En--Es) and English-German (En--De) corpus~\cite{di-gangi-etal-2019-must}. All the datasets include source audios with the corresponding transcriptions in source language and translations in target language. 
For the Augmented Librispeech En--Fr corpus, we follow previous work~\cite{wang-etal-2020-curriculum} and use the 100-hour clean training set with aligned references and provided Google translations, resulting in double size of training pairs. 
For MUST-C datasets, 
We use the official data splits for train and development and tst-COMMON set for test. The statistics for these three datasets are listed in Table~\ref{tab:stat}.



\subsection{Experimental Settings}
We use 80-dimensional log-mel filterbanks as acoustic features, which are calculated with 25ms window size and 10ms step size and normalized by utterance-level Cepstral Mean and Variance Normalization (CMVN). For transcriptions and translations, SentencePiece\footnote{\url{https://github.com/google/sentencepiece}}~\cite{kudo-richardson-2018-sentencepiece} is used to generate subword vocabularies with the sizes of 4k and 8k respectively. We remove the punctuation in transcriptions.
\begin{table}[t]
\small
\begin{center}
\setlength{\tabcolsep}{1.8mm}
\begin{tabular}{l|c c c}
\toprule[1pt]
Corpus&Train&Dev&Test\\
\midrule[0.5pt]
En--Fr&47,271$\times$2 (100h)&1,071 (2.0h)&2,048 (4.0h)\\
En--Es&265,625 (496h)&1,316 (2.5h)&2,502 (4.0h)\\
En--De&229,703 (400h)&1,423 (2.5h)&2,641 (4.0h)\\
\bottomrule[1pt]
\end{tabular} 
\end{center}
\caption{
The number of sentences and the duration of
audios for Augmented LibriSpeech En--Fr, MuST-C En--Es and En--De datasets. 
}
\label{tab:stat}
\end{table}


Our acoustic encoder follows the settings of the original Conv-Transformer \cite{DBLP:conf/interspeech/HuangHYC20}, except that the channel number in convolution layers and hidden size and head number in Transformer layers are half values of theirs. This means the output dimension of the acoustic encoder is 256. We use 6 Transformer layers in the semantic encoder and 4 in the ST decoder. The hyper-parameters $\lambda$ (Eq.~\ref{eq:bp-ctc-loss}), $\mu$ (Eq.~\ref{eq:we-shrink}) and $\alpha$ (Eq.~\ref{eq:loss}) in our model are set to $0.5$, $1.0$ and $1.0$, respectively.

Our model is trained with 8 NVIDIA Tesla V100 GPUs, batched with an approximate 40000-frame features. We use Adam optimizer~\cite{DBLP:journals/corr/KingmaB14} with a 0.002 learning rate and 10000 warm-up steps followed by the inverse square root scheduler. Dropout strategy is used with a rate of $0.1$.
We save checkpoints every epoch and average the last 10 checkpoints for evaluation with a beam size of $5$.
For simplicity, we use the same K and N values as those of training for inference.  We implement our model
based on Fairseq S2T\footnote{\url{https://github.com/pytorch/fairseq/tree/master/examples/speech_to_text}}~\cite{wang-etal-2020-fairseq}.

\subsection{Evaluation Metrics}
We apply SacreBLEU\footnote{\url{https://github.com/mjpost/sacreBLEU}} for translation quality evaluation unless otherwise stated. For the metrics of latency, we adapt Average Proportion (AP)~\cite{DBLP:journals/corr/ChoE16} and Average Lagging (AL)~\cite{ma-etal-2019-stacl} to ST settings, following previous studies~\cite{ren-etal-2020-simulspeech,ma-etal-2020-simulmt}.

\paragraph{Average Proportion.}
AP calculates the mean absolute latency cost
by each target token, where we replace the steps of source tokens with the time spent. It can be calculated as follows:
\vspace{-0.8em}
\begin{equation} \label{eq:ap} \small
     AP(\bm{x},\bm{y}) = \frac{1}{|\bm{x}||\bm{y}|} \sum_{i=1}^{|\bm{y}|} d(y_i)
\vspace{-0.8em}
\end{equation}
where $d(y_i)$ is the speech duration that has been listened when producing the target token $y_i$.

\paragraph{Average Lagging.}
AL evaluates the degree of that the user is out of sync with the speaker, in terms of the number of source tokens~\cite{ma-etal-2019-stacl}. Following \citet{ma-etal-2020-simulmt}, we also extend it to the basis of time duration rather than source tokens, which is defined as follows:
\vspace{-0.8em}
\begin{equation} \label{eq:al} \small
     AL(\bm{x},\bm{y}) = \frac{1}{\tau(|\bm{x}|)} \sum_{i=1}^{\tau(|\bm{x}|)} [d(y_i) - \frac{|\bm{x}|}{|\bm{y}^*|} T_s (i-1)]
\vspace{-0.8em}
\end{equation}
where $|\bm{y}^*|$ is the length of the reference translation, and $\tau(|\bm{x}|)$ denotes the index of the corresponding target token when our model has read the entire source speech. $T_s$ represents that the speech features are extracted every $T_s$ ms (decided by the step size in the feature extraction and downsampling rates in convolution layers), which will be 80ms in our model. What's more, our Conv-Transformer module introduces a 140ms look-ahead window~\cite{DBLP:conf/interspeech/HuangHYC20}, so we add 140 to the final AL scores to be fairly compared with other models.
\begin{figure}[!htb]
\centering
\subfigure[Augmented LibriSpeech En--Fr]{
\includegraphics[width=0.45\textwidth]{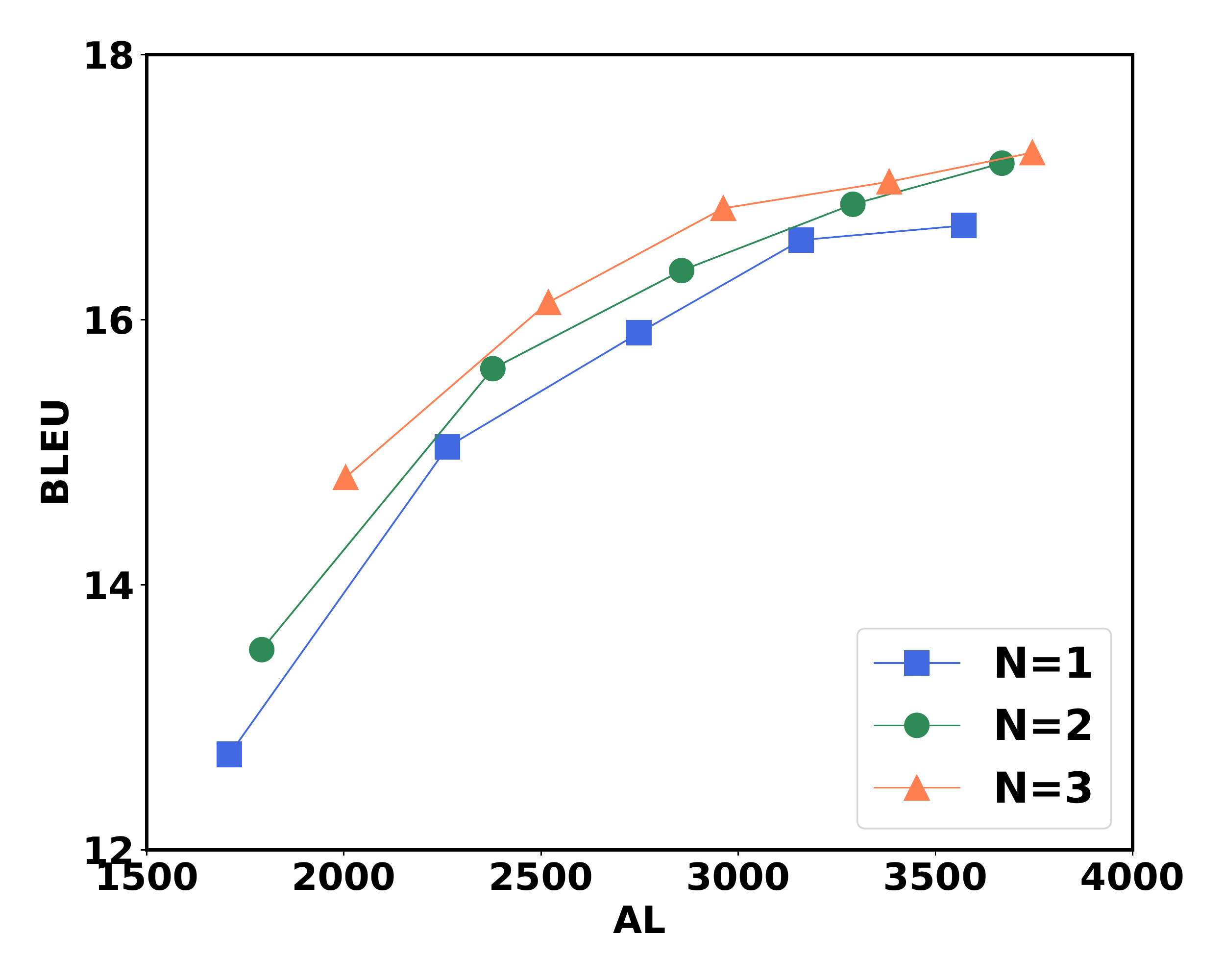}\label{subfig:main-en-fr}
}
\subfigure[MUST-C En--Es]{
\includegraphics[width=0.45\textwidth]{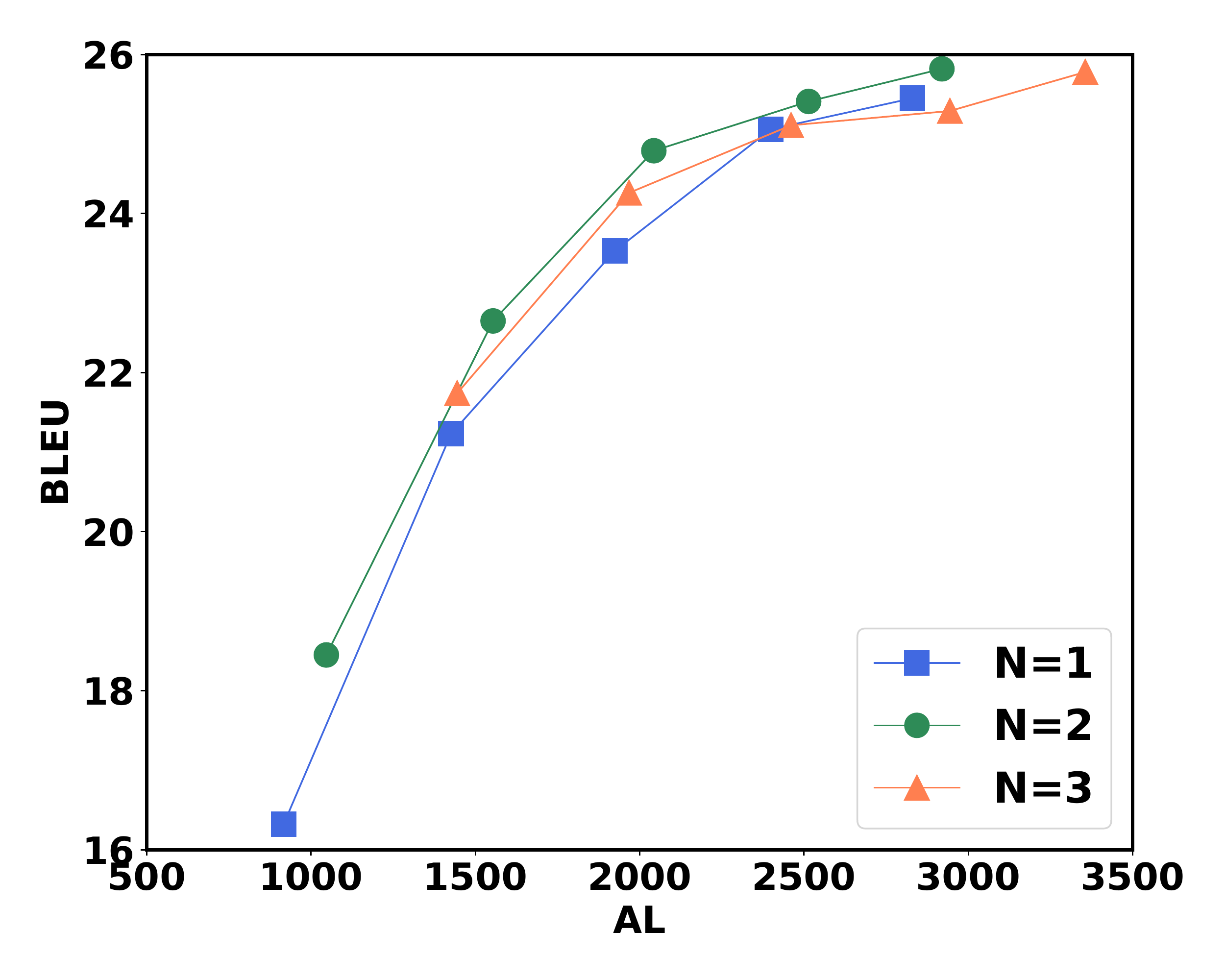}\label{subfig:main-en-es}
}
\subfigure[MUST-C En--De]{
\includegraphics[width=0.45\textwidth]{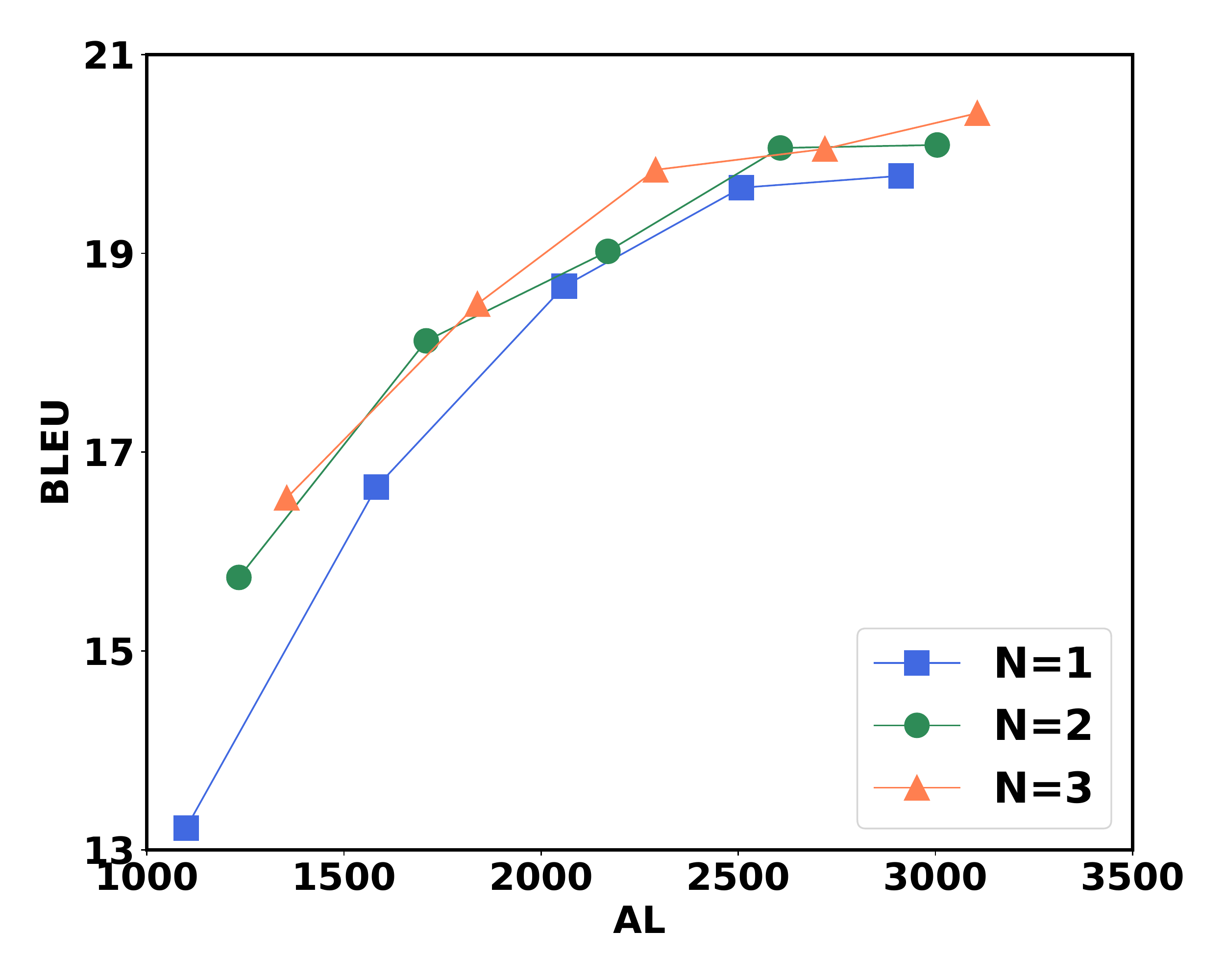}\label{subfig:main-en-de}
}
\caption{
Translation quality (BLEU) vs. latency (AL) of our RealTranS model with Wait-K-Stride-N simultaneous strategy.  For each dataset, we display the results of N$=$1, 2 and 3, with K$=$N, N+2, N+4, N+6, N+8. 
}
\label{fig:main-res}
\end{figure}
\section{Experimental Results}
This section displays our experimental results. To explicitly show the performance trend of models in different scenarios, we use line charts to display most of the compared results. Their corresponding numeric results can be found in Appendix~\ref{sec:appendix}.

\subsection{Translation Quality vs. Latency}
We first evaluate our RealTranS model with our Wait-K-Stride-N simultaneous strategy on the three datasets. We select three values 1, 2 and 3 for N to compare (it becomes conventional Wait-K when N$=$1). The results are displayed in Figure~\ref{fig:main-res}.

Results show that RealTranS achieves higher BLEU scores as the K value increases, with sacrfice of translation delay, consistent with prior works \cite{ren-etal-2020-simulspeech,ma-etal-2020-simulmt}. 

Compared to the conventional Wait-K (N$=$1),
our model with N$=$2 can achieve better BLEU scores under the same latency requirements, which demonstrates the effectiveness of our proposed Wait-K-Stride-N strategy.
When N$=$3, the latency becomes higher. And it only achieves similar gains in BLEU scores compared to N$=$2 on MUST-C En--Es and En--De datasets. Therefore, we will use N$=$2 as our simultaneous strategy in later experiments unless otherwise stated.

\subsection{Comparison with SimulSpeech}
\begin{figure}[t]
\centering
\subfigure[Translation quality vs. latency in terms of AP]{
\includegraphics[width=0.45\textwidth]{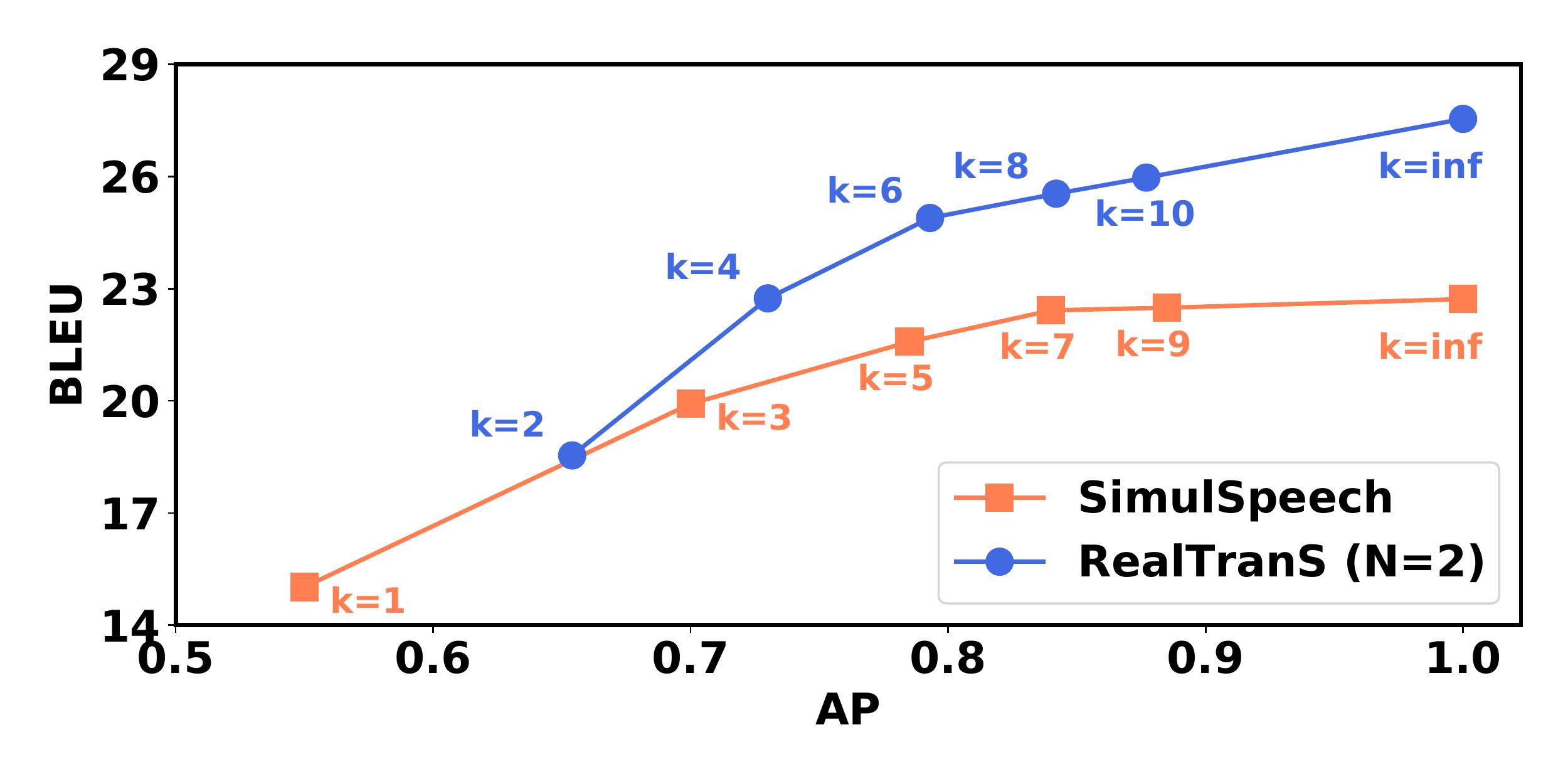}\label{subfig:ss-ap}
}
\subfigure[Translation quality vs. latency in terms of AL]{
\includegraphics[width=0.45\textwidth]{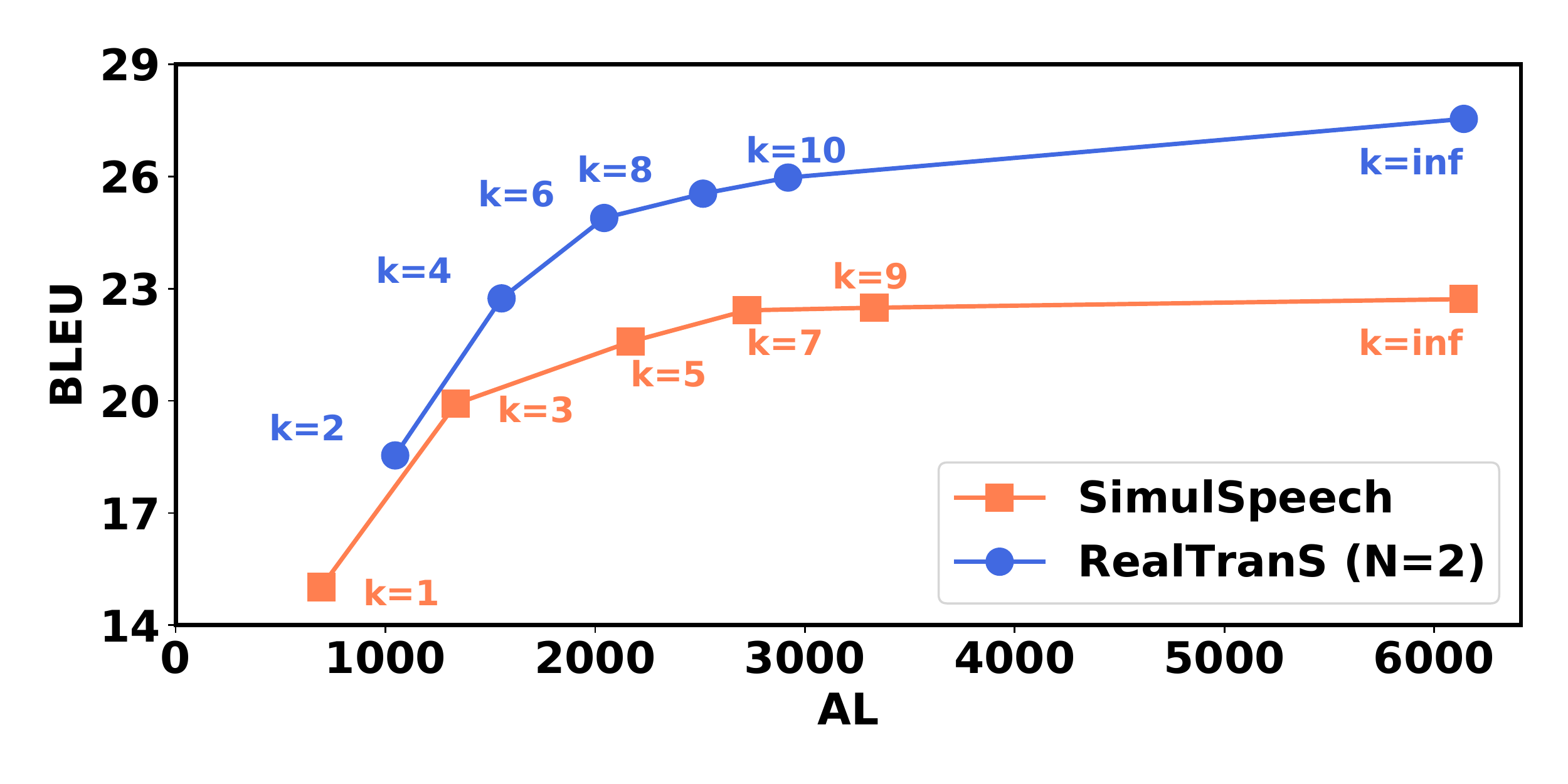}\label{subfig:ss-al}
}
\caption{
Translation quality (BLEU) vs. latency (AP and AL) comparison on MUST-C En-Es dataset.
}
\label{fig:simulspeech}
\end{figure}
We compare with SimulSpeech~\cite{ren-etal-2020-simulspeech}, the state-of-the-art end-to-end model for SST. Figure~\ref{fig:simulspeech} shows the performance comparison on the MUST-C En-Es dataset (we report tokenized case-sensitive BLEU scores following their settings). Since they only report segment-based AL, we transfer it to our time-based AL proportionally based on the latency when K$=$inf.

We find that our RealTranS model outperforms SimulSpeech almost in all latency settings, with an average of about $3$ higher BLEU scores. Although SimulSpeech achieves relatively lower latency (e.g. less than 1000 ms AL when K$=$1), the performance inevitably suffers.
What's more, SimulSpeech has leveraged multi-task learning and knowledge distillation to enhance their performance, which can be also applied to further improve the performance of our RealTranS model.

\subsection{Comparison with Cascaded Model}
We implement a cascaded model to compare with RealTranS under the same latency. Specifically, we combine our acoustic encoder (Section~\ref{sec:mod-ae}) and a Transformer decoder as our ASR model and use the conventional Transformer encoder-decoder architecture as our NMT model. Their configuration is similar to RealTranS (e.g. the same hidden dimension and the same number of Transformer layers). And we train the ASR and NMT model with the same corpus. The conventional Wait-K strategy is used in the ASR model since the alignment between speech and transcription is monotonic, while Wait-K-Stride-N is applied in the NMT model. Since several combinations of the ASR and NMT models may be under the same latency, 
we report the best BLEU score among them.

\begin{table}[t]
\small
\begin{center}
\setlength{\tabcolsep}{1.3mm}
\begin{tabular}{l|c c c c c c}
\toprule[1pt]
Model & K$=$2 & K$=$4 & K$=$6 & K$=$8 & K$=$10 & K$=$inf\\
\midrule[0.5pt]
Cascaded & 14.92 & 19.22 & 22.11 & 23.33 & 24.47 & 26.79 \\
RealTranS & \textbf{18.45} & \textbf{22.65} & \textbf{24.79} & \textbf{25.41} & \textbf{25.82} & \textbf{27.40}\\
\bottomrule[1pt]
\end{tabular} 
\end{center}
\vskip -0.5em
\caption{
Comparison with the cascaded model using Wait-K-Stride-N strategy when N$=$2 in terms of BLEU scores. The same K value means similar latency. }
\vskip -0.5em
\label{tab:cascaded}
\end{table}
Table~\ref{tab:cascaded} shows the comparison results on MUST-C En--Es dataset. We have the following observations: 1) our RealTranS model outperforms the cascaded model in all latency settings, which demonstrates the superiority of RealTranS. 2) The improvement over the cascaded model becomes larger when the value of K is smaller. This observation is consistent with \citet{ren-etal-2020-simulspeech}. We attribute this to the advantage of end-to-end models over cascaded models, where the impact of error propagation in cascaded models may be amplified when the latency is low.

\subsection{Effects of Blank Penalty and Weighted Shrinking Operation}
In this subsection, we examine the effects of our proposed methods, including blank penalty (Eq.~\ref{eq:bp-ctc-loss}) and weighted-shrinking operation (Eq.~\ref{eq:we-shrink}).

\paragraph{Blank Penalty.} We propose a blank penalty to alleviate the inaccuracy of alignments between speech features and transcriptions when applying unidirectional encoders for SST (Section~\ref{sec:mod-ae}). To examine the effect, we evaluate the shrinking quality, i.e., the differences between the length of representations after the shrinking operation and that of the ground-truth transcription, on MUST-C En--Es dataset, and display the statistics in Table~\ref{tab:shrink-quality}.
We can see that the performance, as well as the shrinking quality, drops when removing CTC pre-training. It further decreases when removing the blank penalty, which shows its effectiveness.
Also, we can see that the performance loss partly comes from using unidirectional encoders rather than bidirectional for simultaneous purposes (the $4$th row), which can be compensated by the blank penalty.

\begin{table}[t]
\small
\begin{center}
\setlength{\tabcolsep}{1.0mm}
\begin{tabular}{l| c c c |c }
\toprule[1pt]
Model & Diff$\le$2 & Diff$\le$4 &Diff$\le$6 & BLEU\\
\midrule[0.5pt]
Full Model& \textbf{52\%} & \textbf{74\%} & \textbf{85\%} & \textbf{27.40} \\
$\;\;$ - CTC PT& 48\% & 70\% & 82\% & 26.63 \\
$\;\;\;\;$ - BP & 35\% & 53\% & 67\% & 25.76 \\
$\;\;\;\;\;\;$ + Bi-Enc& 42\% & 59\% & 72\% & 26.58 \\
\bottomrule[1pt]
\end{tabular} 
\end{center}
\caption{
Shrinking quality of RealTranS. ``CTC PT'' indicates that the acoustic encoder is pre-trained with CTC. ``BP'' represents blank penalty, and ``Bi-Enc'' means that using bidirectional Transformer encoders. ``Diff $\le$ $n$'' means the difference between the length of shrunk representations and that of the ground-truth transcription is less than or equal to $n$. We report the percentage of the cases on the MUST-C En--Es test set. The BLEU scores displayed are results when K$=$inf. }
\label{tab:shrink-quality}
\end{table}
\begin{figure}[ht]
\centering
\subfigure[Different shrinking methods]{
\includegraphics[width=0.35\textwidth]{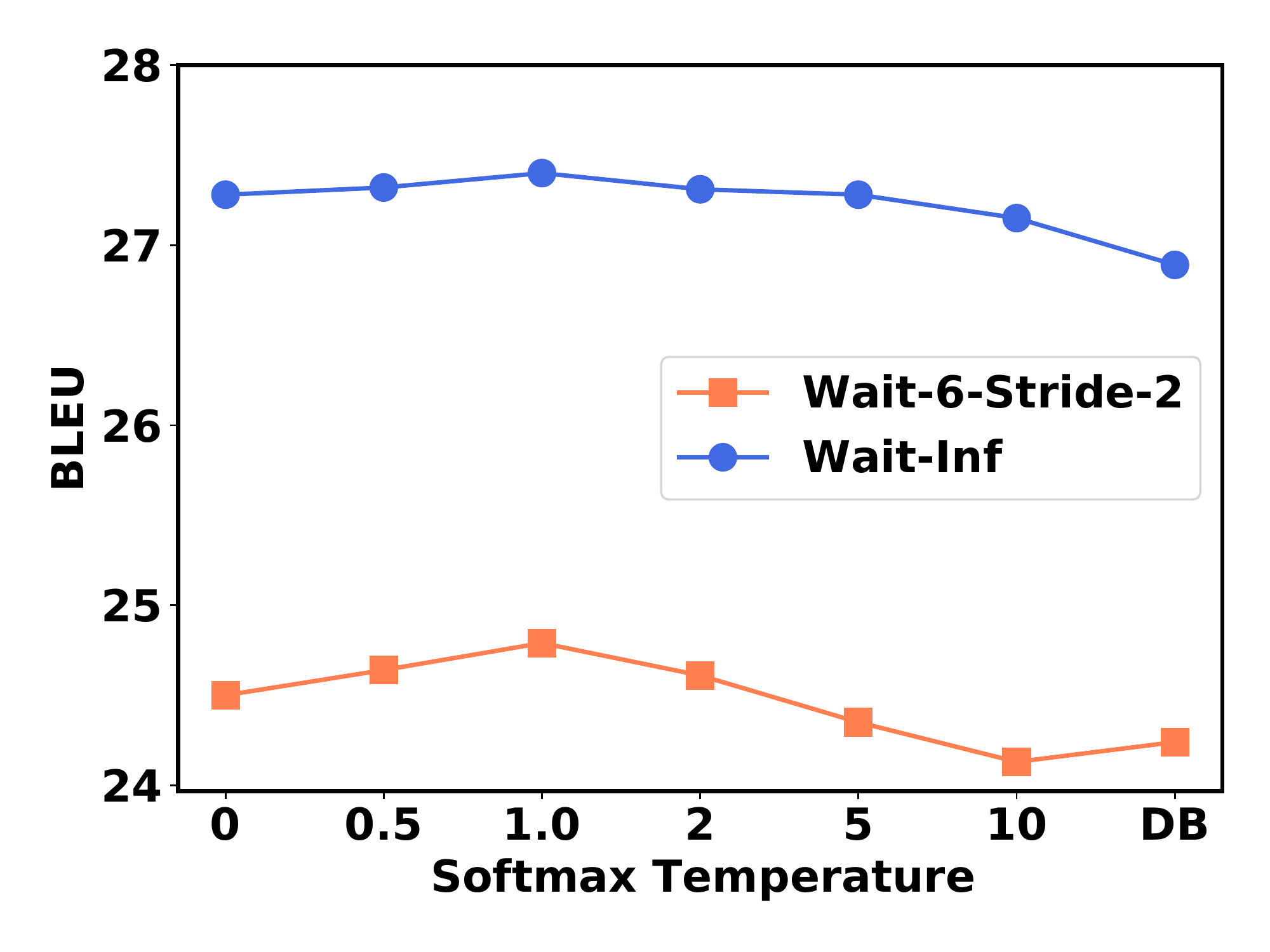}
\label{subfig:sh-temperature}
}
\subfigure[Different downsampling methods]{
\includegraphics[width=0.35\textwidth]{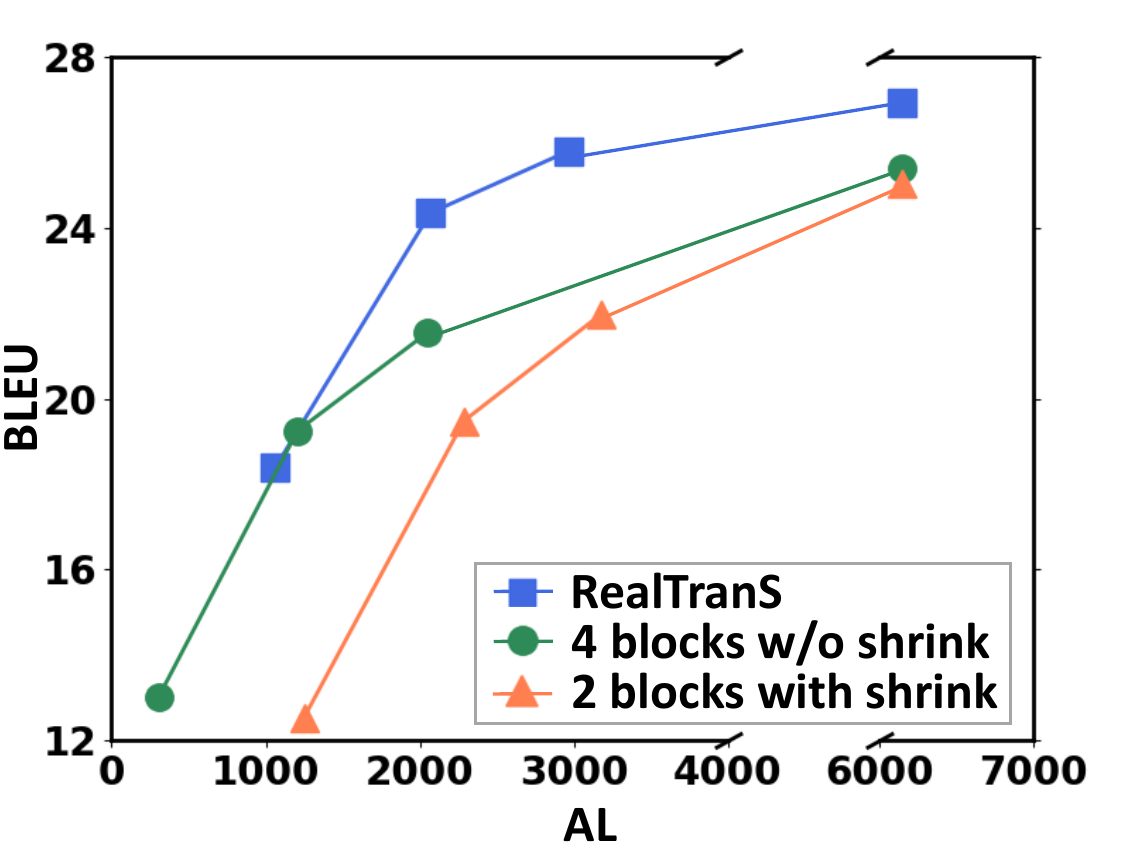}
\label{subfig:sh-blocks}
}
\caption{
Upper: the BLEU scores for different shrinking methods, where ``DB'' means simply dropping blank frames. Lower: BLEU-AL tendency for different downsampling methods, where ``n blocks'' represents using n Conv-Transformer blocks in the acoustic encoder.
}
\label{fig:shrinking}
\end{figure}
\paragraph{Weighted-Shrinking.}
To validate our weighted-shrinking operation, we first investigate the effects of various values for the shrinking temperature $\mu$ in Eq.~\ref{eq:we-shrink} and display the results in Figure~\ref{subfig:sh-temperature}.
The results show that our weighted-shrinking mechanism ($\mu=1.0$) performs better than both simply averaging all the frames ($\mu=0$) and dropping blank frames (``DB'').

We also try to replace our weighted-shrinking module with another Conv-Transformer block (see Section~\ref{sec:mod-ae}), resulting in a model with 240ms downsample rate in total (denoted as ``4 blocks w/o shrink''). Figure~\ref{subfig:sh-blocks} shows comparison results, together with a model with only 2 blocks (40ms downsample rate, denoted as ``2 blocks with shrink''). We can find that downsampling only with convolution layers performs worse than RealTranS, while less downsample rate also affects the performance. This implies that there is an upper-bound for downsampling with only convolution layers while maintaining performance, and our weighted-shrinking operation can be used as an addition to further improve the performance.

\begin{figure}[t]
\centering
\includegraphics[width=0.48\textwidth]{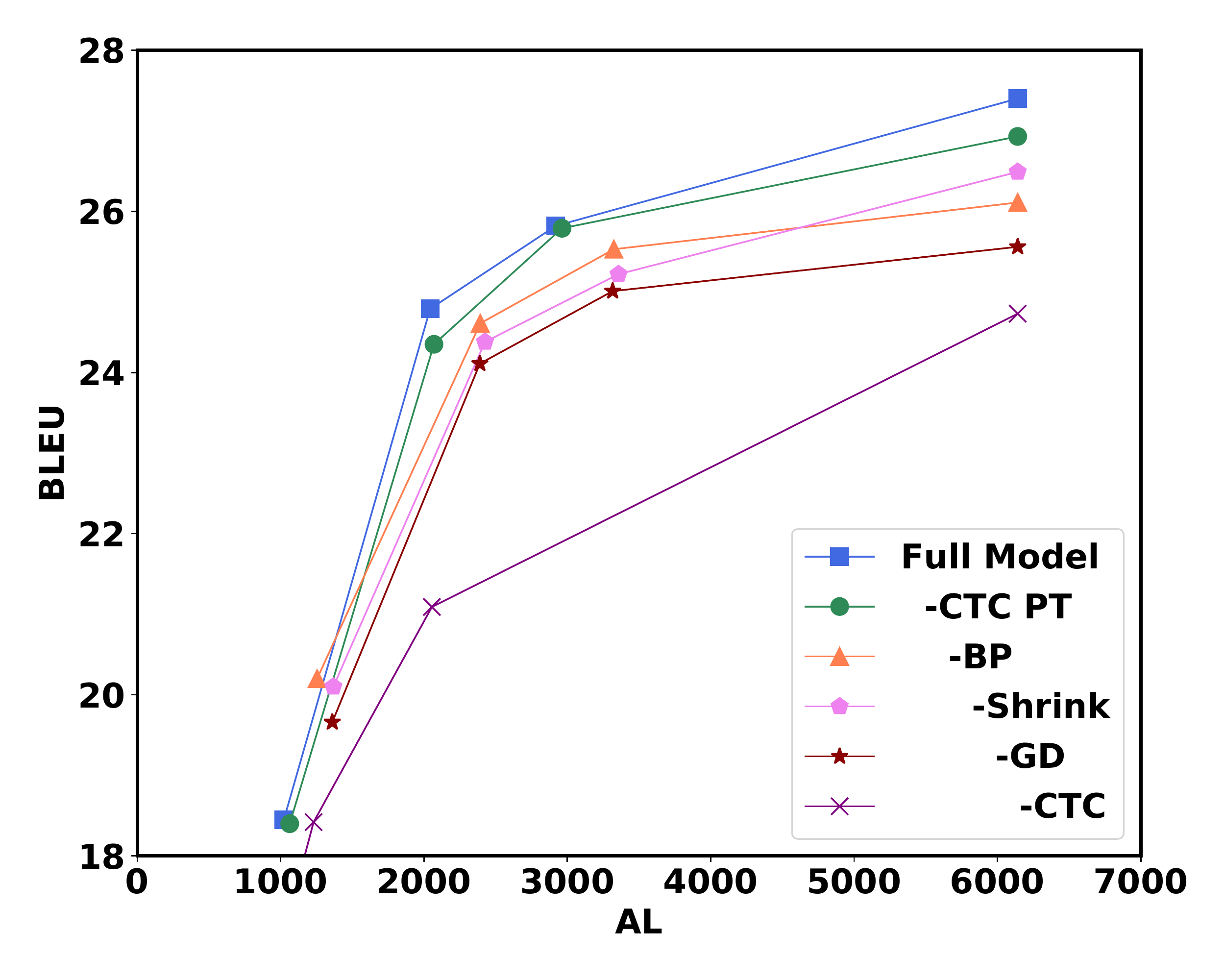}
\vskip -0.5em
\caption{
Ablation study for RealTranS. We choose the results of 3 latency settings and full sentence translation for each ablation. 
``GD'' means gradual downsampling.
}
\vskip -0.5em
\label{fig:ablation}
\end{figure}
\subsection{Ablation Study}
We evaluate the contributions of different modules in RealTranS. Each module is evaluated in four kinds of latency settings: Wait-2-Stride-2, Wait-6-Stride-2, Wait-10-Stride-2, and also Wait-Inf (full sentence translation). The results are shown in Figure~\ref{fig:ablation}, where ``-CTC PT'' means we do not pre-train the encoder with CTC loss and ``-BP'' indicates we further remove the blank penalty. ``-shrink'' means removing the weighted-shrinking operation and the semantic encoder, while ``-GD'' denotes disabling the gradual downsampling by moving the Transformer layers in block 1 \& 2 to block 3 (see Figure~\ref{fig:model}) to perform downsampling at the beginning layers of the acoustic encoder. Finally, ``-CTC'' indicates that CTC objective is removed.

Figure~\ref{fig:ablation} shows that all modules play a role in RealTranS. Specifically, we have the following observations: 1) The CTC module is important for improving translation quality, and it can be further improved by pre-training. 2) The blank penalty is useful in reducing latency while maintaining translation quality. 3) The blank penalty is essential for the shrinking operation since on full sentence translation the shrinking operation degrades the performance without the blank penalty  (``-BP'' vs ``-Shrink''). 4) Gradual downsampling also contributes to the performance, because directly downsampling to a large rate may make it difficult to learn acoustic features.

\subsection{Comparison in Full Sentence Translation}
\begin{table}[t]
\small
\begin{center}
\setlength{\tabcolsep}{1.5mm}
\begin{tabular}{l|l|l }
\toprule[1pt]
Dataset & Method & BLEU\\
\midrule[0.5pt]
En-Fr & Transformer+KD~\cite{DBLP:conf/interspeech/LiuXZHWWZ19} & 17.02 \\
& TCEN-LSTM~\cite{DBLP:conf/aaai/WangWLY020} & 17.05 \\
& Curriculum PT~\cite{wang-etal-2020-curriculum} & 17.66 \\
& LUT~\cite{DBLP:journals/corr/abs-2009-09704} & 17.75 \\ 
& STAST~\cite{DBLP:journals/corr/abs-2010-14920} & 17.81 \\ 
& COSTT~\cite{DBLP:journals/corr/abs-2009-09737} & 17.83 \\
&Transformer+AFS~\cite{zhang-etal-2020-adaptive} & 18.56$^\star$ \\
& RealTranS (ours) & 18.97 \\
&  & 18.30$^\star$\\
\midrule[0.5pt]
En-De & Transformer+MAM~\cite{DBLP:journals/corr/abs-2010-11445} & 21.87$^\star$ \\
& Transformer+ML~\cite{DBLP:journals/corr/abs-1911-04283} & 22.11$^\star$ \\
&Transformer+AFS~\cite{zhang-etal-2020-adaptive} & 22.38$^\star$ \\
&Fairseq S2T~\cite{wang-etal-2020-fairseq} & 22.70$^\star$ \\
&Espnet ST~\cite{inaguma-etal-2020-espnet} & 22.91$^\star$ \\
& STAST~\cite{DBLP:journals/corr/abs-2010-14920} & 23.06 \\ 
& RealTranS (ours) & 23.53\\
&  & 22.99$^\star$\\
\bottomrule[1pt]
\end{tabular} 
\end{center}
\caption{
Full sentence translation results on 
Augmented LibriSpeech En--Fr and 
MUST-C En--De datasets. Results marked with $^\star$ are case-sensitive BLEU scores, while the others are case-insensitive.}
\label{tab:full-sent}
\end{table}
Although focusing on SST, RealTranS can also be applied in full sentence ST. For fair comparison, we replace the unidirectional Transformer layers in RealTranS with bidirectional ones and report both case-insensitive and case-sensitive BLEU scores following prior works. Table~\ref{tab:full-sent} displays the BLEU scores compared with existing methods (we only compare with end-to-end models trained with the same data for fair comparison) on Augmented LibriSpeech En--Fr and MUST-C En--De dataset. RealTranS yields competitive (En--Fr) or even better (En--De) results, even though most of these prior methods depend on some extra techniques like pre-training decoders or using SpecAugment~\cite{DBLP:conf/interspeech/ParkCZCZCL19}. This validates the superiority of our proposed architecture.

\section{Conclusion}
This work proposes a new end-to-end model RealTranS and a new strategy Wait-K-Stride-N for SST.
RealTranS gradually bridges the modality gap between speech and text, and achieves new state-of-the-art results for SST.
Empirical studies have shown the proposed
blank penalty for CTC loss helps on the alignment with transcription, which reduces latency while maintaining translation quality. Our weighted-shrinking operation, as well as Wait-K-Stride-N simultaneous strategy, further improves the performance.
We also compare RealTranS with other methods for full sentence translation, where RealTranS still exhibits competitive results, showing its superiority.

\bibliographystyle{acl_natbib}
\bibliography{anthology,acl2021}

\begin{thebibliography}{49}
\expandafter\ifx\csname natexlab\endcsname\relax\def\natexlab#1{#1}\fi

\bibitem[{Anastasopoulos and Chiang(2018)}]{anastasopoulos-chiang-2018-tied}
Antonios Anastasopoulos and David Chiang. 2018.
\newblock \href {https://doi.org/10.18653/v1/N18-1008} {Tied multitask learning
  for neural speech translation}.
\newblock In \emph{Proceedings of the 2018 Conference of the North {A}merican
  Chapter of the Association for Computational Linguistics: Human Language
  Technologies, Volume 1 (Long Papers)}, pages 82--91, New Orleans, Louisiana.
  Association for Computational Linguistics.

\bibitem[{Arivazhagan et~al.(2019)Arivazhagan, Cherry, Macherey, Chiu, Yavuz,
  Pang, Li, and Raffel}]{arivazhagan-etal-2019-monotonic}
Naveen Arivazhagan, Colin Cherry, Wolfgang Macherey, Chung-Cheng Chiu, Semih
  Yavuz, Ruoming Pang, Wei Li, and Colin Raffel. 2019.
\newblock \href {https://doi.org/10.18653/v1/P19-1126} {Monotonic infinite
  lookback attention for simultaneous machine translation}.
\newblock In \emph{Proceedings of the 57th Annual Meeting of the Association
  for Computational Linguistics}, pages 1313--1323, Florence, Italy.
  Association for Computational Linguistics.

\bibitem[{Bahar et~al.(2021)Bahar, Bieschke, Schl{\"{u}}ter, and
  Ney}]{DBLP:conf/slt/BaharBSN21}
Parnia Bahar, Tobias Bieschke, Ralf Schl{\"{u}}ter, and Hermann Ney. 2021.
\newblock \href {https://doi.org/10.1109/SLT48900.2021.9383462} {Tight
  integrated end-to-end training for cascaded speech translation}.
\newblock In \emph{{IEEE} Spoken Language Technology Workshop, {SLT} 2021,
  Shenzhen, China, January 19-22, 2021}, pages 950--957. {IEEE}.

\bibitem[{Bahar et~al.(2019)Bahar, Zeyer, Schl{\"{u}}ter, and
  Ney}]{DBLP:journals/corr/abs-1911-08876}
Parnia Bahar, Albert Zeyer, Ralf Schl{\"{u}}ter, and Hermann Ney. 2019.
\newblock \href {http://arxiv.org/abs/1911.08876} {On using specaugment for
  end-to-end speech translation}.
\newblock \emph{CoRR}, abs/1911.08876.

\bibitem[{Bansal et~al.(2018)Bansal, Kamper, Livescu, Lopez, and
  Goldwater}]{DBLP:conf/interspeech/BansalKLLG18}
Sameer Bansal, Herman Kamper, Karen Livescu, Adam Lopez, and Sharon Goldwater.
  2018.
\newblock \href {https://doi.org/10.21437/Interspeech.2018-1326} {Low-resource
  speech-to-text translation}.
\newblock In \emph{Interspeech 2018, 19th Annual Conference of the
  International Speech Communication Association, Hyderabad, India, 2-6
  September 2018}, pages 1298--1302. {ISCA}.

\bibitem[{Bansal et~al.(2019)Bansal, Kamper, Livescu, Lopez, and
  Goldwater}]{bansal-etal-2019-pre}
Sameer Bansal, Herman Kamper, Karen Livescu, Adam Lopez, and Sharon Goldwater.
  2019.
\newblock \href {https://doi.org/10.18653/v1/N19-1006} {Pre-training on
  high-resource speech recognition improves low-resource speech-to-text
  translation}.
\newblock In \emph{Proceedings of the 2019 Conference of the North {A}merican
  Chapter of the Association for Computational Linguistics: Human Language
  Technologies, Volume 1 (Long and Short Papers)}, pages 58--68, Minneapolis,
  Minnesota. Association for Computational Linguistics.

\bibitem[{Berard et~al.(2018)Berard, Besacier, Kocabiyikoglu, and
  Pietquin}]{DBLP:conf/icassp/BerardBKP18}
Alexandre Berard, Laurent Besacier, Ali~Can Kocabiyikoglu, and Olivier
  Pietquin. 2018.
\newblock \href {https://doi.org/10.1109/ICASSP.2018.8461690} {End-to-end
  automatic speech translation of audiobooks}.
\newblock In \emph{2018 {IEEE} International Conference on Acoustics, Speech
  and Signal Processing, {ICASSP} 2018, Calgary, AB, Canada, April 15-20,
  2018}, pages 6224--6228. {IEEE}.

\bibitem[{Berard et~al.(2016)Berard, Pietquin, Servan, and
  Besacier}]{DBLP:journals/corr/BerardPSB16}
Alexandre Berard, Olivier Pietquin, Christophe Servan, and Laurent Besacier.
  2016.
\newblock \href {http://arxiv.org/abs/1612.01744} {Listen and translate: {A}
  proof of concept for end-to-end speech-to-text translation}.
\newblock \emph{CoRR}, abs/1612.01744.

\bibitem[{Chen et~al.(2020)Chen, Ma, Zheng, and
  Huang}]{DBLP:journals/corr/abs-2010-11445}
Junkun Chen, Mingbo Ma, Renjie Zheng, and Liang Huang. 2020.
\newblock \href {https://arxiv.org/abs/2010.11445} {{MAM:} masked acoustic
  modeling for end-to-end speech-to-text translation}.
\newblock \emph{CoRR}, abs/2010.11445.

\bibitem[{Cho and Esipova(2016)}]{DBLP:journals/corr/ChoE16}
Kyunghyun Cho and Masha Esipova. 2016.
\newblock \href {http://arxiv.org/abs/1606.02012} {Can neural machine
  translation do simultaneous translation?}
\newblock \emph{CoRR}, abs/1606.02012.

\bibitem[{Dalvi et~al.(2018)Dalvi, Durrani, Sajjad, and
  Vogel}]{dalvi-etal-2018-incremental}
Fahim Dalvi, Nadir Durrani, Hassan Sajjad, and Stephan Vogel. 2018.
\newblock \href {https://doi.org/10.18653/v1/N18-2079} {Incremental decoding
  and training methods for simultaneous translation in neural machine
  translation}.
\newblock In \emph{Proceedings of the 2018 Conference of the North {A}merican
  Chapter of the Association for Computational Linguistics: Human Language
  Technologies, Volume 2 (Short Papers)}, pages 493--499, New Orleans,
  Louisiana. Association for Computational Linguistics.

\bibitem[{Di~Gangi et~al.(2019)Di~Gangi, Cattoni, Bentivogli, Negri, and
  Turchi}]{di-gangi-etal-2019-must}
Mattia~A. Di~Gangi, Roldano Cattoni, Luisa Bentivogli, Matteo Negri, and Marco
  Turchi. 2019.
\newblock \href {https://doi.org/10.18653/v1/N19-1202} {{M}u{ST}-{C}: a
  {M}ultilingual {S}peech {T}ranslation {C}orpus}.
\newblock In \emph{Proceedings of the 2019 Conference of the North {A}merican
  Chapter of the Association for Computational Linguistics: Human Language
  Technologies, Volume 1 (Long and Short Papers)}, pages 2012--2017,
  Minneapolis, Minnesota. Association for Computational Linguistics.

\bibitem[{Dong et~al.(2020{\natexlab{a}})Dong, Wang, Zhou, Xu, Xu, and
  Li}]{DBLP:journals/corr/abs-2009-09737}
Qianqian Dong, Mingxuan Wang, Hao Zhou, Shuang Xu, Bo~Xu, and Lei Li.
  2020{\natexlab{a}}.
\newblock \href {https://arxiv.org/abs/2009.09737} {Consecutive decoding for
  speech-to-text translation}.
\newblock \emph{CoRR}, abs/2009.09737.

\bibitem[{Dong et~al.(2020{\natexlab{b}})Dong, Wang, Zhou, Xu, Xu, and
  Li}]{DBLP:journals/corr/abs-2009-09704}
Qianqian Dong, Mingxuan Wang, Hao Zhou, Shuang Xu, Bo~Xu, and Lei Li.
  2020{\natexlab{b}}.
\newblock \href {https://arxiv.org/abs/2009.09704} {Listen, understand and
  translate: Triple supervision decouples end-to-end speech-to-text
  translation}.
\newblock \emph{CoRR}, abs/2009.09704.

\bibitem[{F{\"{u}}gen et~al.(2007)F{\"{u}}gen, Waibel, and
  Kolss}]{DBLP:journals/mt/FugenWK07}
Christian F{\"{u}}gen, Alex Waibel, and Muntsin Kolss. 2007.
\newblock \href {https://doi.org/10.1007/s10590-008-9047-0} {Simultaneous
  translation of lectures and speeches}.
\newblock \emph{Mach. Transl.}, 21(4):209--252.

\bibitem[{Gangi et~al.(2019)Gangi, Negri, and
  Turchi}]{DBLP:conf/interspeech/GangiNT19}
Mattia Antonino~Di Gangi, Matteo Negri, and Marco Turchi. 2019.
\newblock \href {https://doi.org/10.21437/Interspeech.2019-3045} {Adapting
  transformer to end-to-end spoken language translation}.
\newblock In \emph{Interspeech 2019, 20th Annual Conference of the
  International Speech Communication Association, Graz, Austria, 15-19
  September 2019}, pages 1133--1137. {ISCA}.

\bibitem[{Graves et~al.(2006)Graves, Fern{\'{a}}ndez, Gomez, and
  Schmidhuber}]{DBLP:conf/icml/GravesFGS06}
Alex Graves, Santiago Fern{\'{a}}ndez, Faustino~J. Gomez, and J{\"{u}}rgen
  Schmidhuber. 2006.
\newblock \href {https://doi.org/10.1145/1143844.1143891} {Connectionist
  temporal classification: labelling unsegmented sequence data with recurrent
  neural networks}.
\newblock In \emph{Machine Learning, Proceedings of the Twenty-Third
  International Conference {(ICML} 2006), Pittsburgh, Pennsylvania, USA, June
  25-29, 2006}, volume 148 of \emph{{ACM} International Conference Proceeding
  Series}, pages 369--376. {ACM}.

\bibitem[{Gu et~al.(2017)Gu, Neubig, Cho, and Li}]{gu-etal-2017-learning}
Jiatao Gu, Graham Neubig, Kyunghyun Cho, and Victor~O.K. Li. 2017.
\newblock \href {https://www.aclweb.org/anthology/E17-1099} {Learning to
  translate in real-time with neural machine translation}.
\newblock In \emph{Proceedings of the 15th Conference of the {E}uropean Chapter
  of the Association for Computational Linguistics: Volume 1, Long Papers},
  pages 1053--1062, Valencia, Spain. Association for Computational Linguistics.

\bibitem[{Huang et~al.(2020)Huang, Hu, Yeung, and
  Chen}]{DBLP:conf/interspeech/HuangHYC20}
Wenyong Huang, Wenchao Hu, Yu~Ting Yeung, and Xiao Chen. 2020.
\newblock \href {https://doi.org/10.21437/Interspeech.2020-2361}
  {Conv-transformer transducer: Low latency, low frame rate, streamable
  end-to-end speech recognition}.
\newblock In \emph{Interspeech 2020, 21st Annual Conference of the
  International Speech Communication Association, Virtual Event, Shanghai,
  China, 25-29 October 2020}, pages 5001--5005. {ISCA}.

\bibitem[{Inaguma et~al.(2020)Inaguma, Kiyono, Duh, Karita, Yalta, Hayashi, and
  Watanabe}]{inaguma-etal-2020-espnet}
Hirofumi Inaguma, Shun Kiyono, Kevin Duh, Shigeki Karita, Nelson Yalta, Tomoki
  Hayashi, and Shinji Watanabe. 2020.
\newblock \href {https://doi.org/10.18653/v1/2020.acl-demos.34} {{ESP}net-{ST}:
  All-in-one speech translation toolkit}.
\newblock In \emph{Proceedings of the 58th Annual Meeting of the Association
  for Computational Linguistics: System Demonstrations}, pages 302--311,
  Online. Association for Computational Linguistics.

\bibitem[{Indurthi et~al.(2019)Indurthi, Han, Lakumarapu, Lee, Chung, Kim, and
  Kim}]{DBLP:journals/corr/abs-1911-04283}
Sathish~Reddy Indurthi, Houjeung Han, Nikhil~Kumar Lakumarapu, Beomseok Lee,
  Insoo Chung, Sangha Kim, and Chanwoo Kim. 2019.
\newblock \href {http://arxiv.org/abs/1911.04283} {Data efficient direct
  speech-to-text translation with modality agnostic meta-learning}.
\newblock \emph{CoRR}, abs/1911.04283.

\bibitem[{Iranzo-S{\'a}nchez et~al.(2020)Iranzo-S{\'a}nchez,
  Gim{\'e}nez~Pastor, Silvestre-Cerd{\`a}, Baquero-Arnal, Civera~Saiz, and
  Juan}]{iranzo-sanchez-etal-2020-direct}
Javier Iranzo-S{\'a}nchez, Adri{\`a} Gim{\'e}nez~Pastor, Joan~Albert
  Silvestre-Cerd{\`a}, Pau Baquero-Arnal, Jorge Civera~Saiz, and Alfons Juan.
  2020.
\newblock \href {https://doi.org/10.18653/v1/2020.emnlp-main.206} {Direct
  segmentation models for streaming speech translation}.
\newblock In \emph{Proceedings of the 2020 Conference on Empirical Methods in
  Natural Language Processing (EMNLP)}, pages 2599--2611, Online. Association
  for Computational Linguistics.

\bibitem[{Jia et~al.(2019)Jia, Johnson, Macherey, Weiss, Cao, Chiu, Ari,
  Laurenzo, and Wu}]{DBLP:conf/icassp/JiaJMWCCALW19}
Ye~Jia, Melvin Johnson, Wolfgang Macherey, Ron~J. Weiss, Yuan Cao,
  Chung{-}Cheng Chiu, Naveen Ari, Stella Laurenzo, and Yonghui Wu. 2019.
\newblock \href {https://doi.org/10.1109/ICASSP.2019.8683343} {Leveraging
  weakly supervised data to improve end-to-end speech-to-text translation}.
\newblock In \emph{{IEEE} International Conference on Acoustics, Speech and
  Signal Processing, {ICASSP} 2019, Brighton, United Kingdom, May 12-17, 2019},
  pages 7180--7184. {IEEE}.

\bibitem[{Kingma and Ba(2015)}]{DBLP:journals/corr/KingmaB14}
Diederik~P. Kingma and Jimmy Ba. 2015.
\newblock \href {http://arxiv.org/abs/1412.6980} {Adam: {A} method for
  stochastic optimization}.
\newblock In \emph{3rd International Conference on Learning Representations,
  {ICLR} 2015, San Diego, CA, USA, May 7-9, 2015, Conference Track
  Proceedings}.

\bibitem[{Kocabiyikoglu et~al.(2018)Kocabiyikoglu, Besacier, and
  Kraif}]{DBLP:conf/lrec/KocabiyikogluBK18}
Ali~Can Kocabiyikoglu, Laurent Besacier, and Olivier Kraif. 2018.
\newblock \href
  {http://www.lrec-conf.org/proceedings/lrec2018/summaries/621.html}
  {Augmenting librispeech with french translations: {A} multimodal corpus for
  direct speech translation evaluation}.
\newblock In \emph{Proceedings of the Eleventh International Conference on
  Language Resources and Evaluation, {LREC} 2018, Miyazaki, Japan, May 7-12,
  2018}. European Language Resources Association {(ELRA)}.

\bibitem[{Kudo and Richardson(2018)}]{kudo-richardson-2018-sentencepiece}
Taku Kudo and John Richardson. 2018.
\newblock \href {https://doi.org/10.18653/v1/D18-2012} {{S}entence{P}iece: A
  simple and language independent subword tokenizer and detokenizer for neural
  text processing}.
\newblock In \emph{Proceedings of the 2018 Conference on Empirical Methods in
  Natural Language Processing: System Demonstrations}, pages 66--71, Brussels,
  Belgium. Association for Computational Linguistics.

\bibitem[{Liu et~al.(2018)Liu, Jin, and Zhang}]{DBLP:conf/nips/LiuJZ18}
Hu~Liu, Sheng Jin, and Changshui Zhang. 2018.
\newblock \href
  {http://papers.nips.cc/paper/7363-connectionist-temporal-classification-with-maximum-entropy-regularization}
  {Connectionist temporal classification with maximum entropy regularization}.
\newblock In \emph{Advances in Neural Information Processing Systems 31: Annual
  Conference on Neural Information Processing Systems 2018, NeurIPS 2018,
  December 3-8, 2018, Montr{\'{e}}al, Canada}, pages 839--849.

\bibitem[{Liu et~al.(2019)Liu, Xiong, Zhang, He, Wu, Wang, and
  Zong}]{DBLP:conf/interspeech/LiuXZHWWZ19}
Yuchen Liu, Hao Xiong, Jiajun Zhang, Zhongjun He, Hua Wu, Haifeng Wang, and
  Chengqing Zong. 2019.
\newblock \href {https://doi.org/10.21437/Interspeech.2019-2582} {End-to-end
  speech translation with knowledge distillation}.
\newblock In \emph{Interspeech 2019, 20th Annual Conference of the
  International Speech Communication Association, Graz, Austria, 15-19
  September 2019}, pages 1128--1132. {ISCA}.

\bibitem[{Liu et~al.(2020)Liu, Zhu, Zhang, and
  Zong}]{DBLP:journals/corr/abs-2010-14920}
Yuchen Liu, Junnan Zhu, Jiajun Zhang, and Chengqing Zong. 2020.
\newblock \href {https://arxiv.org/abs/2010.14920} {Bridging the modality gap
  for speech-to-text translation}.
\newblock \emph{CoRR}, abs/2010.14920.

\bibitem[{Ma et~al.(2019)Ma, Huang, Xiong, Zheng, Liu, Zheng, Zhang, He, Liu,
  Li, Wu, and Wang}]{ma-etal-2019-stacl}
Mingbo Ma, Liang Huang, Hao Xiong, Renjie Zheng, Kaibo Liu, Baigong Zheng,
  Chuanqiang Zhang, Zhongjun He, Hairong Liu, Xing Li, Hua Wu, and Haifeng
  Wang. 2019.
\newblock \href {https://doi.org/10.18653/v1/P19-1289} {{STACL}: Simultaneous
  translation with implicit anticipation and controllable latency using
  prefix-to-prefix framework}.
\newblock In \emph{Proceedings of the 57th Annual Meeting of the Association
  for Computational Linguistics}, pages 3025--3036, Florence, Italy.
  Association for Computational Linguistics.

\bibitem[{Ma et~al.(2020{\natexlab{a}})Ma, Pino, and
  Koehn}]{ma-etal-2020-simulmt}
Xutai Ma, Juan Pino, and Philipp Koehn. 2020{\natexlab{a}}.
\newblock \href {https://www.aclweb.org/anthology/2020.aacl-main.58}
  {{S}imul{MT} to {S}imul{ST}: Adapting simultaneous text translation to
  end-to-end simultaneous speech translation}.
\newblock In \emph{Proceedings of the 1st Conference of the Asia-Pacific
  Chapter of the Association for Computational Linguistics and the 10th
  International Joint Conference on Natural Language Processing}, pages
  582--587, Suzhou, China. Association for Computational Linguistics.

\bibitem[{Ma et~al.(2020{\natexlab{b}})Ma, Wang, Dousti, Koehn, and
  Pino}]{DBLP:journals/corr/abs-2011-00033}
Xutai Ma, Yongqiang Wang, Mohammad~Javad Dousti, Philipp Koehn, and Juan Pino.
  2020{\natexlab{b}}.
\newblock \href {https://arxiv.org/abs/2011.00033} {Streaming simultaneous
  speech translation with augmented memory transformer}.
\newblock \emph{CoRR}, abs/2011.00033.

\bibitem[{Mathias and Byrne(2006)}]{DBLP:conf/icassp/MathiasB06}
Lambert Mathias and William Byrne. 2006.
\newblock \href {https://doi.org/10.1109/ICASSP.2006.1660082} {Statistical
  phrase-based speech translation}.
\newblock In \emph{2006 {IEEE} International Conference on Acoustics Speech and
  Signal Processing, {ICASSP} 2006, Toulouse, France, May 14-19, 2006}, pages
  561--564. {IEEE}.

\bibitem[{Ney(1999)}]{DBLP:conf/icassp/Ney99}
Hermann Ney. 1999.
\newblock \href {https://doi.org/10.1109/ICASSP.1999.758176} {Speech
  translation: coupling of recognition and translation}.
\newblock In \emph{Proceedings of the 1999 {IEEE} International Conference on
  Acoustics, Speech, and Signal Processing, {ICASSP} '99, Phoenix, Arizona,
  USA, March 15-19, 1999}, pages 517--520. {IEEE} Computer Society.

\bibitem[{Oda et~al.(2014)Oda, Neubig, Sakti, Toda, and
  Nakamura}]{oda-etal-2014-optimizing}
Yusuke Oda, Graham Neubig, Sakriani Sakti, Tomoki Toda, and Satoshi Nakamura.
  2014.
\newblock \href {https://doi.org/10.3115/v1/P14-2090} {Optimizing segmentation
  strategies for simultaneous speech translation}.
\newblock In \emph{Proceedings of the 52nd Annual Meeting of the Association
  for Computational Linguistics (Volume 2: Short Papers)}, pages 551--556,
  Baltimore, Maryland. Association for Computational Linguistics.

\bibitem[{Park et~al.(2019)Park, Chan, Zhang, Chiu, Zoph, Cubuk, and
  Le}]{DBLP:conf/interspeech/ParkCZCZCL19}
Daniel~S. Park, William Chan, Yu~Zhang, Chung{-}Cheng Chiu, Barret Zoph,
  Ekin~D. Cubuk, and Quoc~V. Le. 2019.
\newblock \href {https://doi.org/10.21437/Interspeech.2019-2680} {Specaugment:
  {A} simple data augmentation method for automatic speech recognition}.
\newblock In \emph{Interspeech 2019, 20th Annual Conference of the
  International Speech Communication Association, Graz, Austria, 15-19
  September 2019}, pages 2613--2617. {ISCA}.

\bibitem[{Ren et~al.(2020)Ren, Liu, Tan, Zhang, Qin, Zhao, and
  Liu}]{ren-etal-2020-simulspeech}
Yi~Ren, Jinglin Liu, Xu~Tan, Chen Zhang, Tao Qin, Zhou Zhao, and Tie-Yan Liu.
  2020.
\newblock \href {https://doi.org/10.18653/v1/2020.acl-main.350}
  {{S}imul{S}peech: End-to-end simultaneous speech to text translation}.
\newblock In \emph{Proceedings of the 58th Annual Meeting of the Association
  for Computational Linguistics}, pages 3787--3796, Online. Association for
  Computational Linguistics.

\bibitem[{Salesky and Black(2020)}]{salesky-black-2020-phone}
Elizabeth Salesky and Alan~W Black. 2020.
\newblock \href {https://doi.org/10.18653/v1/2020.acl-main.217} {Phone features
  improve speech translation}.
\newblock In \emph{Proceedings of the 58th Annual Meeting of the Association
  for Computational Linguistics}, pages 2388--2397, Online. Association for
  Computational Linguistics.

\bibitem[{Sperber et~al.(2017)Sperber, Neubig, Niehues, and
  Waibel}]{sperber-etal-2017-neural}
Matthias Sperber, Graham Neubig, Jan Niehues, and Alex Waibel. 2017.
\newblock \href {https://doi.org/10.18653/v1/D17-1145} {Neural
  lattice-to-sequence models for uncertain inputs}.
\newblock In \emph{Proceedings of the 2017 Conference on Empirical Methods in
  Natural Language Processing}, pages 1380--1389, Copenhagen, Denmark.
  Association for Computational Linguistics.

\bibitem[{Stoian et~al.(2020)Stoian, Bansal, and
  Goldwater}]{DBLP:conf/icassp/StoianBG20}
Mihaela~C. Stoian, Sameer Bansal, and Sharon Goldwater. 2020.
\newblock \href {https://doi.org/10.1109/ICASSP40776.2020.9053847} {Analyzing
  {ASR} pretraining for low-resource speech-to-text translation}.
\newblock In \emph{2020 {IEEE} International Conference on Acoustics, Speech
  and Signal Processing, {ICASSP} 2020, Barcelona, Spain, May 4-8, 2020}, pages
  7909--7913. {IEEE}.

\bibitem[{Wang et~al.(2020{\natexlab{a}})Wang, Tang, Ma, Wu, Okhonko, and
  Pino}]{wang-etal-2020-fairseq}
Changhan Wang, Yun Tang, Xutai Ma, Anne Wu, Dmytro Okhonko, and Juan Pino.
  2020{\natexlab{a}}.
\newblock \href {https://www.aclweb.org/anthology/2020.aacl-demo.6} {Fairseq
  {S}2{T}: Fast speech-to-text modeling with fairseq}.
\newblock In \emph{Proceedings of the 1st Conference of the Asia-Pacific
  Chapter of the Association for Computational Linguistics and the 10th
  International Joint Conference on Natural Language Processing: System
  Demonstrations}, pages 33--39, Suzhou, China. Association for Computational
  Linguistics.

\bibitem[{Wang et~al.(2020{\natexlab{b}})Wang, Wu, Liu, Yang, and
  Zhou}]{DBLP:conf/aaai/WangWLY020}
Chengyi Wang, Yu~Wu, Shujie Liu, Zhenglu Yang, and Ming Zhou.
  2020{\natexlab{b}}.
\newblock \href {https://aaai.org/ojs/index.php/AAAI/article/view/6452}
  {Bridging the gap between pre-training and fine-tuning for end-to-end speech
  translation}.
\newblock In \emph{The Thirty-Fourth {AAAI} Conference on Artificial
  Intelligence, {AAAI} 2020, The Thirty-Second Innovative Applications of
  Artificial Intelligence Conference, {IAAI} 2020, The Tenth {AAAI} Symposium
  on Educational Advances in Artificial Intelligence, {EAAI} 2020, New York,
  NY, USA, February 7-12, 2020}, pages 9161--9168. {AAAI} Press.

\bibitem[{Wang et~al.(2020{\natexlab{c}})Wang, Wu, Liu, Zhou, and
  Yang}]{wang-etal-2020-curriculum}
Chengyi Wang, Yu~Wu, Shujie Liu, Ming Zhou, and Zhenglu Yang.
  2020{\natexlab{c}}.
\newblock \href {https://doi.org/10.18653/v1/2020.acl-main.344} {Curriculum
  pre-training for end-to-end speech translation}.
\newblock In \emph{Proceedings of the 58th Annual Meeting of the Association
  for Computational Linguistics}, pages 3728--3738, Online. Association for
  Computational Linguistics.

\bibitem[{Weiss et~al.(2017)Weiss, Chorowski, Jaitly, Wu, and
  Chen}]{DBLP:journals/corr/WeissCJWC17}
Ron~J. Weiss, Jan Chorowski, Navdeep Jaitly, Yonghui Wu, and Zhifeng Chen.
  2017.
\newblock \href {http://arxiv.org/abs/1703.08581} {Sequence-to-sequence models
  can directly transcribe foreign speech}.
\newblock \emph{CoRR}, abs/1703.08581.

\bibitem[{Zhang et~al.(2020)Zhang, Titov, Haddow, and
  Sennrich}]{zhang-etal-2020-adaptive}
Biao Zhang, Ivan Titov, Barry Haddow, and Rico Sennrich. 2020.
\newblock \href {https://doi.org/10.18653/v1/2020.findings-emnlp.230} {Adaptive
  feature selection for end-to-end speech translation}.
\newblock In \emph{Findings of the Association for Computational Linguistics:
  EMNLP 2020}, pages 2533--2544, Online. Association for Computational
  Linguistics.

\bibitem[{Zheng et~al.(2020)Zheng, Liu, Zheng, Ma, Liu, and
  Huang}]{zheng-etal-2020-simultaneous}
Baigong Zheng, Kaibo Liu, Renjie Zheng, Mingbo Ma, Hairong Liu, and Liang
  Huang. 2020.
\newblock \href {https://doi.org/10.18653/v1/2020.acl-main.254} {Simultaneous
  translation policies: From fixed to adaptive}.
\newblock In \emph{Proceedings of the 58th Annual Meeting of the Association
  for Computational Linguistics}, pages 2847--2853, Online. Association for
  Computational Linguistics.

\bibitem[{Zheng et~al.(2019{\natexlab{a}})Zheng, Zheng, Ma, and
  Huang}]{zheng-etal-2019-simpler}
Baigong Zheng, Renjie Zheng, Mingbo Ma, and Liang Huang. 2019{\natexlab{a}}.
\newblock \href {https://doi.org/10.18653/v1/D19-1137} {Simpler and faster
  learning of adaptive policies for simultaneous translation}.
\newblock In \emph{Proceedings of the 2019 Conference on Empirical Methods in
  Natural Language Processing and the 9th International Joint Conference on
  Natural Language Processing (EMNLP-IJCNLP)}, pages 1349--1354, Hong Kong,
  China. Association for Computational Linguistics.

\bibitem[{Zheng et~al.(2019{\natexlab{b}})Zheng, Zheng, Ma, and
  Huang}]{zheng-etal-2019-simultaneous}
Baigong Zheng, Renjie Zheng, Mingbo Ma, and Liang Huang. 2019{\natexlab{b}}.
\newblock \href {https://doi.org/10.18653/v1/P19-1582} {Simultaneous
  translation with flexible policy via restricted imitation learning}.
\newblock In \emph{Proceedings of the 57th Annual Meeting of the Association
  for Computational Linguistics}, pages 5816--5822, Florence, Italy.
  Association for Computational Linguistics.

\bibitem[{Zheng et~al.(2019{\natexlab{c}})Zheng, Ma, Zheng, and
  Huang}]{zheng-etal-2019-speculative}
Renjie Zheng, Mingbo Ma, Baigong Zheng, and Liang Huang. 2019{\natexlab{c}}.
\newblock \href {https://doi.org/10.18653/v1/D19-1144} {Speculative beam search
  for simultaneous translation}.
\newblock In \emph{Proceedings of the 2019 Conference on Empirical Methods in
  Natural Language Processing and the 9th International Joint Conference on
  Natural Language Processing (EMNLP-IJCNLP)}, pages 1395--1402, Hong Kong,
  China. Association for Computational Linguistics.

\end{thebibliography}

\clearpage
\appendix
\section*{Appendix}
\section{Numeric Results for the Figures}
\label{sec:appendix}

\begin{table}[H]
\small
\begin{center}
\setlength{\tabcolsep}{2.5mm}
\renewcommand\arraystretch{1.5}
\begin{tabular}{l|c c|c c|c c|c c|c c}
\toprule[1pt]
\multirow{2}{*}{\bf{N Value}}&\multicolumn{2}{c|}{\bf{K=N}}&\multicolumn{2}{c|}{\bf{K=N+2}}&\multicolumn{2}{c|}{\bf{K=N+4}}&\multicolumn{2}{c|}{\bf{K=N+6}}&\multicolumn{2}{c}{\bf{K=N+8}}\\
&\bf{BLEU}&\bf{AL}&\bf{BLEU}&\bf{AL}&\bf{BLEU}&\bf{AL}&\bf{BLEU}&\bf{AL}&\bf{BLEU}&\bf{AL}\\
\midrule[0.5pt]
\bf{N=1}&12.72&1710&15.04&2263&15.9&2748&16.6&3160&16.71&3572\\
\bf{N=2}&13.51&1792&15.63&2378&16.37&2856&16.87&3291&17.18&3669\\
\bf{N=3}&14.81&2004&16.13&2519&16.84&2962&17.04&3383&17.26&3746\\
\bottomrule[1pt]
\end{tabular} 
\end{center}
\caption{
Numeric Results for Figure~\ref{subfig:main-en-fr}. }
\end{table}

\begin{table}[H]
\small
\begin{center}
\setlength{\tabcolsep}{2.5mm}
\renewcommand\arraystretch{1.5}
\begin{tabular}{l|c c|c c|c c|c c|c c}
\toprule[1pt]
\multirow{2}{*}{\bf{N Value}}&\multicolumn{2}{c|}{\bf{K=N}}&\multicolumn{2}{c|}{\bf{K=N+2}}&\multicolumn{2}{c|}{\bf{K=N+4}}&\multicolumn{2}{c|}{\bf{K=N+6}}&\multicolumn{2}{c}{\bf{K=N+8}}\\
&\bf{BLEU}&\bf{AL}&\bf{BLEU}&\bf{AL}&\bf{BLEU}&\bf{AL}&\bf{BLEU}&\bf{AL}&\bf{BLEU}&\bf{AL}\\
\midrule[0.5pt]
\bf{N=1}&16.32&916&21.23&1426&23.53&1924&25.06&2399&25.45&2830\\
\bf{N=2}&18.45&1047&22.65&1554&24.79&2043&25.41&2514&25.82&2920\\
\bf{N=3}&21.74&1445&24.26&1968&25.11&2461&25.29&2944&25.78&3356\\
\bottomrule[1pt]
\end{tabular} 
\end{center}
\caption{
Numeric Results for Figure~\ref{subfig:main-en-es}. }
\end{table}

\begin{table}[H]
\small
\begin{center}
\setlength{\tabcolsep}{2.5mm}
\renewcommand\arraystretch{1.5}
\begin{tabular}{l|c c|c c|c c|c c|c c}
\toprule[1pt]
\multirow{2}{*}{\bf{N Value}}&\multicolumn{2}{c|}{\bf{K=N}}&\multicolumn{2}{c|}{\bf{K=N+2}}&\multicolumn{2}{c|}{\bf{K=N+4}}&\multicolumn{2}{c|}{\bf{K=N+6}}&\multicolumn{2}{c}{\bf{K=N+8}}\\
&\bf{BLEU}&\bf{AL}&\bf{BLEU}&\bf{AL}&\bf{BLEU}&\bf{AL}&\bf{BLEU}&\bf{AL}&\bf{BLEU}&\bf{AL}\\
\midrule[0.5pt]
\bf{N=1}&13.22&1100&16.65&1582&18.67&2059&19.66&2508&19.78&2913\\
\bf{N=2}&15.74&1233&18.12&1709&19.02&2169&20.06&2607&20.09&3005\\
\bf{N=3}&16.54&1355&18.49&1838&19.84&2290&20.05&2720&20.41&3106\\
\bottomrule[1pt]
\end{tabular} 
\end{center}
\caption{
Numeric Results for Figure~\ref{subfig:main-en-de}. }
\end{table}

\begin{table}[H]
\small
\begin{center}
\setlength{\tabcolsep}{1.7mm}
\renewcommand\arraystretch{1.5}
\begin{tabular}{l|c c|c c|c c|c c|c c|c c}
\toprule[1pt]
\multirow{2}{*}{\bf{Model}}&\multicolumn{2}{c|}{\bf{K=N}}&\multicolumn{2}{c|}{\bf{K=N+2}}&\multicolumn{2}{c|}{\bf{K=N+4}}&\multicolumn{2}{c|}{\bf{K=N+6}}&\multicolumn{2}{c|}{\bf{K=N+8}}&\multicolumn{2}{c}{\bf{K=inf}}\\
&\bf{BLEU}&\bf{AP}&\bf{BLEU}&\bf{AP}&\bf{BLEU}&\bf{AP}&\bf{BLEU}&\bf{AP}&\bf{BLEU}&\bf{AP}&\bf{BLEU}&\bf{AP}\\
\midrule[0.5pt]
\bf{SimulSpeech}&15.02&0.550&19.92&0.700&21.58&0.785&22.42&0.840&22.49&0.885&22.72&1.0\\
\bf{RealTranS(N=2)}&18.54&0.654&22.74&0.730&24.89&0.793&25.54&0.842&25.97&0.877&27.54&1.0\\
\bottomrule[1pt]
\end{tabular} 
\end{center}
\caption{
Numeric Results for Figure~\ref{subfig:ss-ap}. }
\end{table}

\begin{table}[H]
\small
\begin{center}
\setlength{\tabcolsep}{1.74mm}
\renewcommand\arraystretch{1.5}
\begin{tabular}{l|c c|c c|c c|c c|c c|c c}
\toprule[1pt]
\multirow{2}{*}{\bf{Model}}&\multicolumn{2}{c|}{\bf{K=N}}&\multicolumn{2}{c|}{\bf{K=N+2}}&\multicolumn{2}{c|}{\bf{K=N+4}}&\multicolumn{2}{c|}{\bf{K=N+6}}&\multicolumn{2}{c|}{\bf{K=N+8}}&\multicolumn{2}{c}{\bf{K=inf}}\\
&\bf{BLEU}&\bf{AL}&\bf{BLEU}&\bf{AL}&\bf{BLEU}&\bf{AL}&\bf{BLEU}&\bf{AL}&\bf{BLEU}&\bf{AL}&\bf{BLEU}&\bf{AL}\\
\midrule[0.5pt]
\bf{SimulSpeech}&15.02&694&19.92&1336&21.58&2169&22.42&2724&22.49&3331&22.72&6141\\
\bf{RealTranS(N=2)}&18.54&1047&22.74&1554&24.89&2043&25.54&2514&25.97&2920&27.54&6141\\
\bottomrule[1pt]
\end{tabular} 
\end{center}
\caption{
Numeric Results for Figure~\ref{subfig:ss-al}. }
\end{table}

\clearpage

\begin{table}[H]
\small
\begin{center}
\setlength{\tabcolsep}{2.0mm}
\renewcommand\arraystretch{1.5}
\begin{tabular}{l|c|c|c|c|c|c|c}
\toprule[1pt]
\bf{Model}&\bf{$\mu$=0}&\bf{$\mu$=0.5}&\bf{$\mu$=1.0}&\bf{$\mu$=2.0}&\bf{$\mu$=5.0}&\bf{$\mu$=10}&\bf{DB}\\
\midrule[0.5pt]
\bf{Wait-6-Stride-2}&24.50&24.64&24.79&24.61&24.35&24.13&24.24\\
\bf{Wait-Inf}&27.28&27.32&27.40&27.31&27.28&27.15&26.89\\
\bottomrule[1pt]
\end{tabular} 
\end{center}
\caption{
Numeric Results for Figure~\ref{subfig:sh-temperature}. }
\end{table}

\begin{table}[H]
\small
\begin{center}
\setlength{\tabcolsep}{2.0mm}
\renewcommand\arraystretch{1.5}
\begin{tabular}{l|c c|c c|c c|c c}
\toprule[1pt]
\multirow{2}{*}{\bf{Model}}&\multicolumn{2}{c|}{\bf{K=2}}&\multicolumn{2}{c|}{\bf{K=6}}&\multicolumn{2}{c|}{\bf{K=10}}&\multicolumn{2}{c}{\bf{K=inf}}\\
&\bf{BLEU}&\bf{AL}&\bf{BLEU}&\bf{AL}&\bf{BLEU}&\bf{AL}&\bf{BLEU}&\bf{AL}\\
\midrule[0.5pt]
\bf{RealTranS}&18.40&1064&24.35&2070&25.79&2963&26.93&6141\\
\bf{4 blocks w/o shrink}&13.03&307&19.25&1207&21.55&2051&25.38&6141\\
\bf{2 blocks with shrink}&12.53&1247&19.48&2286&21.97&3175&25.04&6141\\
\bottomrule[1pt]
\end{tabular} 
\end{center}
\caption{
Numeric Results for Figure~\ref{subfig:sh-blocks}. }
\end{table}

\begin{table}[H]
\small
\begin{center}
\setlength{\tabcolsep}{2.2mm}
\renewcommand\arraystretch{1.5}
\begin{tabular}{l|c c|c c|c c|c c}
\toprule[1pt]
\multirow{2}{*}{\bf{Model}}&\multicolumn{2}{c|}{\bf{K=2}}&\multicolumn{2}{c|}{\bf{K=6}}&\multicolumn{2}{c|}{\bf{K=10}}&\multicolumn{2}{c}{\bf{K=inf}}\\
&\bf{BLEU}&\bf{AL}&\bf{BLEU}&\bf{AL}&\bf{BLEU}&\bf{AL}&\bf{BLEU}&\bf{AL}\\
\midrule[0.5pt]
\bf{Full Model}&18.45&1023&24.79&2043&25.82&2920&27.40&6141\\
\bf{\;\;-CTC PT}&18.40&1064&24.35&2070&25.79&2963&26.93&6141\\
\bf{\;\;\;\;-BP}&20.20&1255&24.61&2393&25.53&3325&26.11&6141\\
\bf{\;\;\;\;\;\;-Shrink}&20.10&1368&24.38&2426&25.22&3357&26.49&6141\\
\bf{\;\;\;\;\;\;\;\;-GD}&19.66&1360&24.11&2390&25.01&3316&25.56&6141\\
\bf{\;\;\;\;\;\;\;\;\;\;-CTC}&12.48&342&18.42&1230&21.09&2055&24.73&6141\\

\bottomrule[1pt]
\end{tabular} 
\end{center}
\caption{
Numeric Results for Figure~\ref{fig:ablation}. }
\end{table}

\end{document}